 \documentclass[letterpaper, 10 pt, conference]{ieeetran}  % Comment this line out if you need a4paper

 \IEEEoverridecommandlockouts                              % This command is only needed if 
\usepackage{graphicx}
\usepackage{subfigure}
\usepackage{booktabs} % for professional tables
\usepackage{caption}
\usepackage[table]{xcolor} 
\usepackage{comment}
\usepackage{multirow}
\usepackage{graphicx}
\usepackage{wrapfig}
\usepackage{paralist}
\usepackage{titlesec}
\usepackage{amsmath}
\usepackage{bbm}
\usepackage[flushmargin]{footmisc}

\usepackage{reledmac}
\usepackage{setspace}
\usepackage{etoolbox}
\usepackage{soul}
\usepackage{amssymb}
\usepackage{makecell}
\usepackage{pifont}% http://ctan.org/pkg/pifont
\usepackage{hyperref}
\usepackage{dsfont}
\usepackage{placeins}

\usepackage[noend]{algpseudocode}

\usepackage{paralist}

\newcommand{\cmark}{\ding{51}}%
\definecolor{orange}{RGB}{179, 98, 0}
  
\definecolor{red}{RGB}{255, 0, 0}
\definecolor{brown}{RGB}{200, 25, 10}
\definecolor{blue}{RGB}{0, 0,255}
\definecolor{green}{RGB}{78, 196, 164}
\definecolor{greenml}{RGB}{0, 200, 250}

\makeatletter
\usepackage{algorithm,algpseudocode,setspace}

\makeatother
\let\footnote\footnoteA

\def\resulttable#1#2{
    \begin{minipage}{.49\textwidth}
    \centering
    \resizebox{\textwidth}{!}{
  \rowcolors{6}{gray!10}{white}
    \begin{tabular}{|l|l|l|l|l|l|l|l|l|}
    \hline
    \multicolumn{1}{|c|}{\multirow{3}{*}{Models}} & \multicolumn{8}{c|}{Dataset  #1} \\ \cline{2-9}
    \multicolumn{1}{|c|}{} & \multicolumn{1}{c|}{\multirow{2}{*}{\begin{tabular}[c]{@{}c@{}} $|\mathcal{V}|$ \end{tabular}}} & \multicolumn{1}{c|}{\multirow{2}{*}{\begin{tabular}[c]{@{}c@{}} $h_c$\end{tabular}}} &  \multicolumn{1}{c|}{\multirow{2}{*}{\begin{tabular}[c]{@{}c@{}} $c_c$\end{tabular}}} & \multicolumn{1}{c|}{\multirow{2}{*}{\begin{tabular}[c]{@{}c@{}} $s_c$ \\ \end{tabular}}} &
     \multicolumn{1}{c|}{\multirow{2}{*}{\begin{tabular}[c]{@{}c@{}} $|\mathcal{E}|$\end{tabular}}} &
     \multicolumn{1}{c|}{\multirow{2}{*}{\begin{tabular}[c]{@{}c@{}} $c_e$ \end{tabular}}} & \multicolumn{2}{c|}{Paths scores} 
     \\ \cline{8-9} 
    \multicolumn{1}{|c|}{} & \multicolumn{1}{c|}{}  & \multicolumn{1}{c|}{} & \multicolumn{1}{c|}{} & \multicolumn{1}{c|}{}   & \multicolumn{1}{c|}{} & \multicolumn{1}{c|}{} & \multicolumn{1}{c|}{\textit{\% all}} & \multicolumn{1}{c|}{\textit{\% any}}   \\ \hline
    #2
    \end{tabular}%
    }
    \end{minipage}
}

\usepackage{hyperref}

\title{\LARGE \bf
     Comparing Reconstruction- and Contrastive-based Models \\
 for Visual Task Planning
}

\author{Constantinos Chamzas*$^{1}$, Martina Lippi*$^{2}$, Michael C. Welle*$^{3}$,\\ Anastasia Varava$^{3}$, Lydia E. Kavraki$^{1}$, and Danica Kragic$^{3}$% <-this % stops a space
\thanks{*Contributed equally (listed in alphabetical order) }% <-this % stops a space
\thanks{This project has been supported in part by NSF GRFP 1842494 (CC) and NSF 2008720 (LEK)}
\thanks{$^{1}$ Rice University,
        6100 Main St, Houston, TX 77005, USA
        {\tt\small chamzas, kavraki@rice.edu}}%
\thanks{$^{2}$Roma Tre University,
        Via Ostiense, 133B, 00154 Roma RM, Italy
        {\tt\small martina.lippi@uniroma3.it}}%
\thanks{$^{3}$KTH Royal Institute of Technology,
        Brinellvägen 8, 114 28 Stockholm, Sweden
        {\tt\small mwelle, varava, dani@kth.se}}%
}

\begin{document}

\maketitle
\thispagestyle{empty}
\pagestyle{empty}

%%%%%%%%%%%%%%%%%%%%%%%%%%%%%%%%%%%%%%%%%%%%%%%%%%%%%%%%%%%%%%%%%%%%%%%%%%%%%%%%
\begin{abstract}

Learning state representations enables robotic planning directly from raw observations such as images.
Most methods learn state representations by utilizing losses based on the reconstruction of the raw observations from a lower-dimensional latent space.
The similarity between observations in the space of images is often assumed and used as a proxy for estimating similarity between the underlying
states of the system.
However, observations commonly contain task-irrelevant factors of variation which are nonetheless important for reconstruction, such as varying lighting and different camera viewpoints.
In this work, we define relevant evaluation metrics and perform a thorough study of different loss functions for state representation learning.
We show that models exploiting  task priors, such as Siamese networks with a simple contrastive loss, outperform reconstruction-based representations in visual  task planning.

\end{abstract}

%%%%%%%%%%%%%%%%%%%%%%%%%%%%%%%%%%%%%%%%%%%%%%%%%%%%%%%%%%%%%%%%%%%%%%%%%%%%%%%%

\section{Introduction}
\begin{figure}[t]
            \centering
            \includegraphics[width= 0.95\linewidth]{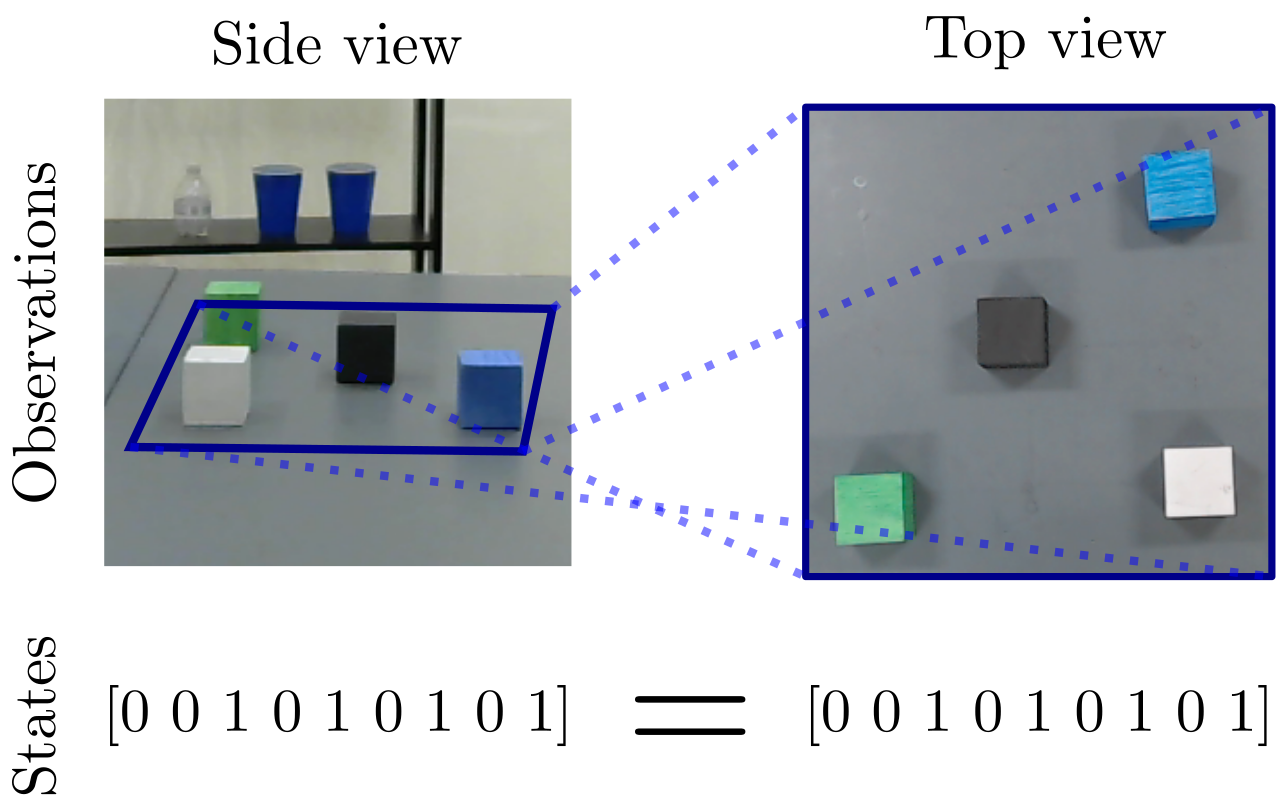}
            \caption{Examples of visually different observations (different views) of the same state (arrangement of the boxes). }
            \label{fig:example_obs_vs_state}
            %\vspace{-\baselineskip}
        \end{figure}

% {\red TO DO: discuss about task priors: are they assumptions? CC: I would say yes they are, but we don't have to adress this explicitly. \\ }
Learning  of low-dimensional state representations from high-dimensional observations such as images have gained significant attention in robotics \cite{konidaris2018skills, stooke2020decoupling}.  
For complex manipulation planning tasks,  this approach is a viable alternative since analytic approaches may be computationally expensive or impossible to define. %{\blue For instance, in this work we consider a box manipulation task on a real robotic system as well as a simulated box manipulation and shelf arrangement task. }
Existing approaches are generally based on an implicit assumption that \emph{similar observations, close in the image space, correspond to similar system states}. 
However, the same underlying state may be related to very different observations due to other factors of variation, such as different views or background of the scene, see \autoref{fig:example_obs_vs_state}.  This is especially true in \emph{task} planning, which we focus on, where states are typically discrete and their  observations may be captured in very different time intervals, leading to the natural occurrence of task irrelevant factors of variation. Similar considerations also hold for task and motion planning (TAMP) settings \cite{dantam2016incremental}. %has been shown above. 
%In this work, we focus on task planning where the discrete nature of the planning is more prone  
%{similar} {\red ML: I would not add similar -  I wanted to say that observations can be taken in different times so theere may be irrelevant objects that appear that lead to dissimilar observations (with similar states)}

To address this, it is crucial to identify the \emph{task-relevant} factors of variation. 
 A step in this direction is done by~\cite{thomas2018disentangling}, where an agent interacts with the environment and tries to disentangle the controllable factors of variation. 
However, if data is being collected in realistic scenarios, irrelevant factors of variation may occur that are difficult to control.

Although several solutions exist in literature, 
%What is missing in the existing literature is 
a unified analysis of representation losses and their influence to the performance of learned representations  for high-level visual task planning is currently missing.  In this work, we perform a systematic comparison of different representation learning methods which can possibly leverage task priors in quasi-static tasks. 
 %\cc{I think we should change our motivation a bit and say it is an analysis to actually proposing our representation, which is learning discrete states in qusi-static tasks leveraging self-supervised priors?. The reviewers are actually treating this work like that anyway}
 To this aim, we also design and collect datasets where the underlying states of the system do not uniquely correspond to observations (images). We study a box manipulation task on a real robotic system as well as a simulated shelf arrangement task. %{\red ML: decide for box stacking - we could cite the workshop for the box?}
In all tasks, 
 different task-irrelevant factors, such as different viewpoints of the same scene or ``distractor'' objects, are present. 
 %\begin{wrapfigure}{rt}{0.4\textwidth}
    %\vspace{-\baselineskip}
   % \end{wrapfigure}
 Our work 
makes the following contributions:
\emph{i)} We introduce evaluation metrics and provide a systematic study for assessing the effect of different loss functions on state representation. Robotic tasks on both real hardware and simulation are analyzed. 
\emph{ii)} We examine a simple data augmentation procedure for contrastive-based models.
 \emph{iii)} We show how task priors in contrastive-based models combined with simple data augmentations can lead to the best performance in  visual task planning with task-irrelevant factors of variation and 
demonstrate the performance of the best derived representations on a real-world robotic task. 
\emph{iv)} We create and distribute datasets for comparing state representation models\footnote{\label{fn:website}{ \mbox{\url{https://state-representation.github.io/web/}}}}.
\section{Related Work}

State representation learning from high-dimensional data has been successfully used in a variety of robotic tasks. 
As shown in \autoref{tab:rw_overview}, the used loss functions are usually a combination of the reconstruction, Kullback–Leibler (KL)-divergence, and contrastive loss functions.  
A common approach to use learned state representations is through
learned forward dynamic models as in  \cite{Ichter2019b, watter2015embed, hafner2019learning, pertsch2020long}. These dynamic models predict future observations (images) and are trained to minimize the pixel distance between the observed image and the decoded predicted observation. Among these works, \cite{Ichter2019b} also exploits a KL loss to regularize the latent space. 
Future rewards and actions are predicted instead in \cite{hafner2019dream}, and the image reconstruction loss is solely used to regularize. 
Since in many cases predicting full images is not practical, some approaches attempt to remove task-irrelevant information from the predicted images. In \cite{nair2020goal}, the residual of goal and the current state is reconstructed which contains more relevant information comparing to a raw image. Similarly, in \cite{finn2016deep}  images are transformed through specialized layers that enhance spatial features, such as object locations.
Learned representations leveraging  reconstruction loss have also been used in specific robotic applications, such as \cite{hoque2020visuospatial} for fabric manipulation and \cite{hoof2016stable} for pendulum swing up. 
%For example, in \cite{hoque2020visuospatial}, image flows are predicted for fabric manipulation. 
%{\red In \cite{hoof2016stable} a pendulum is swung up using high-dimensional tactile images. }

As shown in \autoref{tab:rw_overview}, all the aforementioned methods rely on the reconstruction loss. 
However, in many real scenarios, full images might contain redundant information, making the reconstruction loss not applicable. Inspired by the revival of contrastive methods in computer vision \cite{reviewCRL2020}, some recent works rely on contrastive losses to learn efficient state representations. The works in~\cite{stooke2020decoupling, srinivas2020curl, kipf2019contrastive} augment pixel frames through transformations and use a forward momentum encoder to generate positive and negative examples. These examples are then exploited to learn state representations directly in the latent space without the need for a decoder. In \cite{tcn2018}, a purely contrastive loss is used to learn robotic states from video demonstrations where states that are temporally close are considered similar. % {\red same as slowness? }.
In addition, the authors of \cite{zhang2020learning} remove task-irrelevant information by adding distractors during simulation and considering such states similar in their  contrastive loss formulation.
  Contrastive-like losses have also been formulated using task or robotic priors \cite{jonschkowski2015learning} such as slowness \cite{wiskott2002slow}. The latter has been applied in reinforcement learning \cite{legenstein2010reinforcement} with visual observations, and humanoid robotics state representation \cite{hofer2010using}. A no-action/action prior was also used in our previous work \cite{lsriros}, which was used to formulate a combined reconstruction, KL, and contrastive loss. Here, we  leverage the same task prior of \cite{lsriros} as explained in \autoref{sec:problem}. %\autoref{subsec:training}.

The vast majority of the aforementioned methods are concerned with continuous control tasks, whereas in this work we are focusing on quasi-static states tailored towards long-horizon high-level planning \cite{konidaris2018skills}.  In detail, we take representative models employing different loss functions and perform a thorough study by analyzing their performance in robotic planning tasks with and without task priors. 
Such discrete state representations have been learned in literature by object-centric or compositional models like \cite{kipf2019contrastive, jetchevground2013}, however we do not assume any structural relations between observations.

%In this work, {\red In summary, } we take representative models employing different loss functions and perform a thorough study by analyzing their performance in robotic planning tasks with and without task priors. 

%\begin{wraptable}{rt}{0.45\textwidth}
%\vspace{-\baselineskip}
    \begin{table}[ht!]
    \resizebox{0.95\linewidth}{!}{
    \begin{tabular}%{|l|ccc|}
    { m{0.25\textwidth} | {c}m{0.05\columnwidth}  {c}m{0.05\columnwidth}| {c}m{0.05\columnwidth} |  }
    \hline
    Related works & \multicolumn{1}{c}{\begin{tabular}[c]{@{}c@{}}Recon.\end{tabular}} & \multicolumn{1}{c}{\begin{tabular}[c]{@{}c@{}}KL\end{tabular}} & \begin{tabular}[c]{@{}c@{}}Contr.\end{tabular} \\ \hline
     \cite{watter2015embed, hafner2019learning, pertsch2020long,  hafner2019dream, hoque2020visuospatial, hoof2016stable} & {\centering}\cmark &  &  \\ \hline
    \cite{Ichter2019b,nair2020goal, finn2016deep, finn2017deep} & \cmark & \cmark &  \\ \hline
    \cite{stooke2020decoupling, srinivas2020curl, kipf2019contrastive, tcn2018, zhang2020learning}  &  &  & \cmark  \\ \hline
      \cite{lsriros} & \cmark & \cmark & \cmark \\ \hline
    \end{tabular}
    }
%\vspace{-2pt}
    \caption{Overview of loss functions (reconstruction, KL divergence and contrastive) used by state-of-the-art methods. }
    \label{tab:rw_overview}
    %\vspace{-\baselineskip}
    \end{table}
%\end{wraptable}

\begin{figure*}
\centering
\begin{minipage}{.66\textwidth}
 \centering
 \includegraphics[width=.9\linewidth]{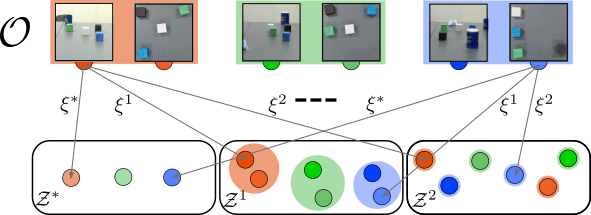}
% \vspace{-2pt}
 \captionof{figure}{ 
    Examples of mapping functions $\xi^*,\xi^1,\xi^2$ (arrows) from observation space $\mathcal{O}$ (top row) to latent spaces $\mathcal{Z}^*, \mathcal{Z}^1, \mathcal{Z}^2$ (bottom row). 
    Boxes arrangement represents the system state  and images marked with variations of the same color contain the same state.}
 \label{fig:problem_explained}
\end{minipage}%
\hspace{4pt}
\begin{minipage}{.001\textwidth}

\end{minipage}
\begin{minipage}{.29\textwidth}
 \centering
 \includegraphics[width=0.65\linewidth]{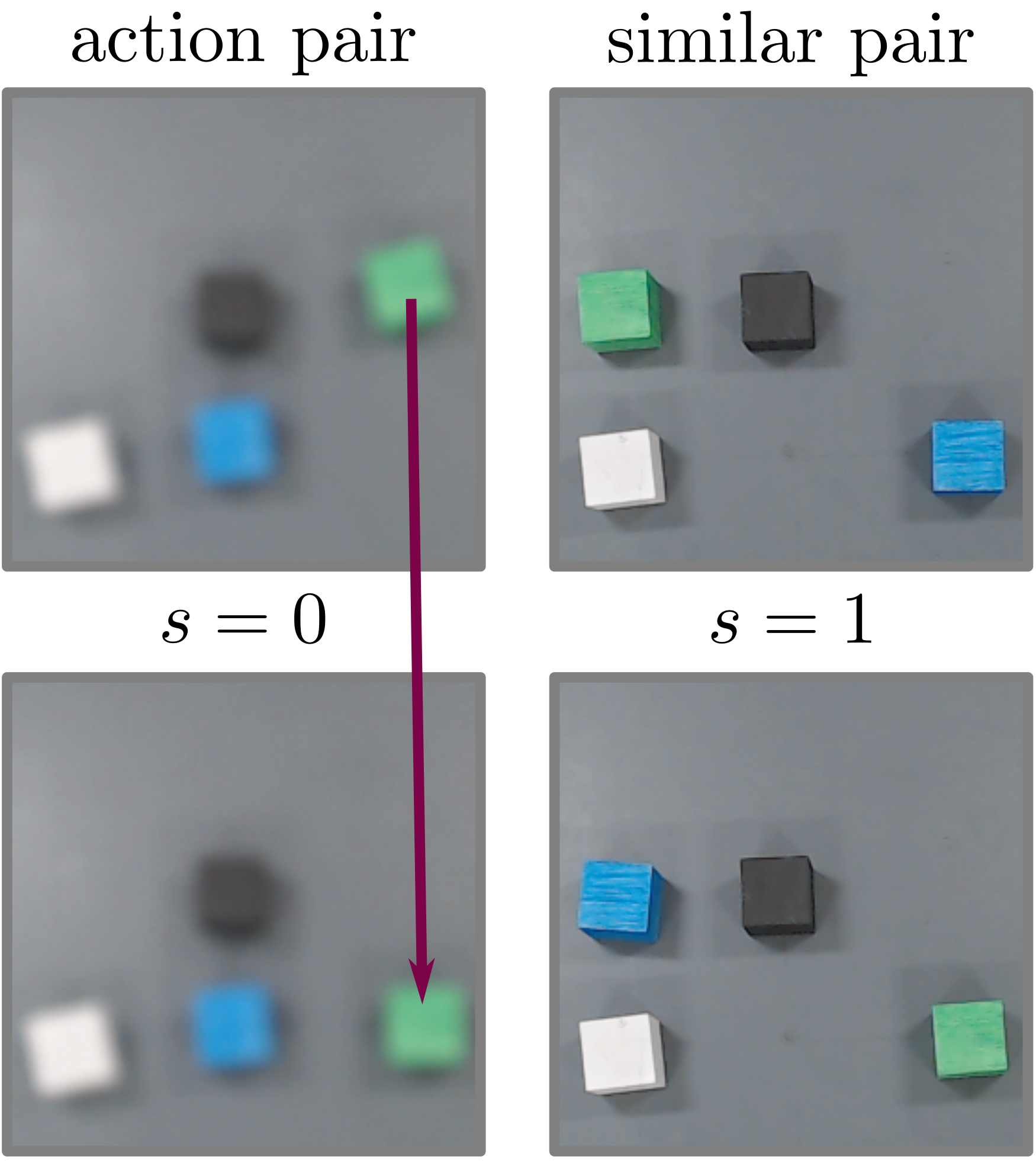}
 %\vspace{-8pt}
 \captionof{figure}{
  Example of action  (left) and similar  (right) pairs. We consider the boxes interchangeable (only the resulting arrangement matters).
 %Examples of action (left) and no-action (right) pairs for a shelf arrangement task (objects are interchangeable). 
 }
 \label{fig:shelf_data_examples}
\end{minipage}
%\vspace{-\baselineskip}
%\vspace{-\baselineskip}
%\vspace{-2pt}
\end{figure*}

\section{Problem Formulation}\label{sec:problem}
 Our objective is to define appropriate state representations for visual task planning given high-dimensional observations provided as images. %{\red ML: list of assumptions? States are discrete? }
 Let $\mathcal{O}$ be the observation space and $\mathcal{Z}$ be a low-dimensional state space, also referred to as latent space. The goal is to define a mapping function \mbox{$\xi:\mathcal{O}\rightarrow \mathcal{Z}$} 
 which extracts the low-dimensional representation $z\in\mathcal{Z} $ given a high-dimensional observation  $o\in\mathcal{O}$. We consider that task-irrelevant factors can be present in the observations which cause them to be  possibly \emph{visually dissimilar} even if they contain the same underlying states.

An ideal mapping function $\xi^*$ should be able to perfectly capture the underlying states of the system despite possible task-irrelevant factors. This means that, given two observations $o_i$ and $o_j$ containing the same state, it holds  $\xi^*(o_i)=\xi^*(o_j)$, i.e., they are mapped into the same latent encoding. We aim to understand how to model $\xi$ such that it is as close as possible to $\xi^*$ when task-irrelevant factors are present in $\mathcal{O}$. 
 
 Although a perfect mapping $\xi^*$ might not be achievable, a good approximation should be able to properly structure the latent space such that encodings associated with the same states are close by, while encodings that are associated with different states are well separated in the latent space.  This implies that the encodings should be \emph{clustered} in the latent space such that each cluster is associated with a possible underlying state of the system. Note that if such clustering is achieved, task planning  can be  easily solved by properly connecting the clusters when an action is allowed between the respective states of the system. Therefore, better mapping results in improved clusters  and requires an easier planning algorithm.  An illustrative example is provided in~\autoref{fig:problem_explained}, where three latent spaces $\mathcal{Z}^*, \mathcal{Z}^1, \mathcal{Z}^2$, obtained with different mapping functions, $\xi^*,\xi^1,\xi^2$, are shown. 
 Considering that observations (top row) in the same colored box contain the same underlying state,  it can be observed that \emph{i)} the latent space $\mathcal{Z}^*$ (bottom left) is optimal since observations containing the same states are mapped exactly into the same latent encoding, \emph{ii)} a sub-optimal latent space $\mathcal{Z}^1$ (bottom middle) is obtained since the latent encodings are properly clustered according to the states of the system, \emph{iii)}  a very hard-to-use latent space $\mathcal{Z}^2$ (bottom right) is obtained where the encodings are not structured according to the  states.

%************************************************************************************************************************
 
 %\subsection{Training Dataset}
 \noindent

\textbf{Training Dataset:} To model the mapping function, we assume  task priors are  available to build the training dataset. In detail, a training dataset $\mathcal{T}_o$ is composed of tuples $(o_i, o_j, s)$ where $o_i, o_j \in \mathcal{O}$ are observations of the system, and \mbox{$s \in \{0, 1\}$} is a signal, obtained from task priors, specifying whether the two observations are 
 \emph{similar} ($s=1$), i.e., they correspond to the same state and $\xi^*(o_i) = \xi^*(o_j)$, or  whether an \emph{action} occurred between them ($s=0$), i.e., they represent consecutive states, implying  that $o_i$ and $o_j$ are dissimilar and  $\xi^*(o_i) \neq \xi^*(o_j)$. An action represent any high-level operation as in \cite{konidaris2018skills}, e.g.,  pick and place,  pushing, and pouring operations. 
We refer to the tuple as a \emph{similar pair} when $s = 1$, and as an \textit{action pair} when $s = 0$.   In addition, the encoded training dataset composed of tuples $(z_i,z_j,s)$, with $z_i=\xi(o_i)$ and $z_j=\xi(o_j)$, is denoted  by $\mathcal{T}_z$.

Note that in both similar and action pairs task-irrelevant factors can change in the observations $o_i,o_j$, i.e., it generally holds $o_i\neq o_j$, while task-relevant factors only change through actions in action pairs. 
%the actions in action pairs modify task-relevant factors, e.g., objects of interest, in the observations but task-irrelevant ones may still change. In contrast, we assume that no change occurs on  task-relevant factors in similar pairs. 
  Moreover,  no knowledge of the underlying states of the training observations is assumed. 
Examples of action and similar pairs are shown in~\autoref{fig:shelf_data_examples} for a box manipulation task (with interchangeable boxes), as detailed in~\autoref{sec:val-setting}. 
The training dataset can be generally collected in self-supervised manner. Indeed, action pairs can be obtained by randomly performing high-level actions with the environment similar to \cite{konidaris2018skills} and recording the respective consecutive observations. Regarding similar pairs, they can be obtained, for example, by recording observations in the tuple  with a certain time separation, leading to the occurrence of different lighting conditions and/or the presence of further irrelevant objects in the scene, or, as in our experiments, by swapping objects if they are interchangable for the task.% at hand. 
%{\red long sentence t the end? }

\textbf{Data Augmentation:}
Inspired by the training procedure in~\cite{chen2020simclear}, we consider a synthetic procedure to generate an additional training dataset $\overline{\mathcal{T}}_o$ from $\mathcal{T}_o$. Let $\mathcal{O}_T$ be the set of all observations  in $\mathcal{T}_o$. The basic idea is that by \emph{randomly} sampling pairs of observations in the dataset, they will \emph{likely} be dissimilar. Therefore, $\overline{\mathcal{T}}_o$ is first initialized to ${\mathcal{T}}_o$. Then, 
for each  similar
%{\blue no-action \st{dissimilar}} 
pair $(o_i, o_j, s = 1)\in\mathcal{T}_o$, we randomly sample $n$ observations $\{o_{1}^s,..., o_{n}^s\}\subset \mathcal{O}_T$ in the dataset and define the tuples $(o_i, o_{k}^s, s = 0),\,k = 1,..,n$, which are added to $\overline{\mathcal{T}}_o$. 
In this way, for each  similar
%no-action 
pair, $n$ novel tuples are introduced in $\overline{\mathcal{T}}_o$ with respect  to ${\mathcal{T}}_o$.  
We experimentally validate that this procedure allows to improve the latent mapping despite possible erroneous tuples in $\overline{\mathcal{T}}_o$, i.e., novel tuples for which it holds $\xi^*(o_i) = \xi^*(o_{k}^s)$.

 \section{Latent Mapping Modeling}\label{sec:models}

 %We demonstrate that the use of  task priors  significantly improves the   representation performance over models employing reconstruction-based losses. This is especially true when task-irrelevant factors of variations are present in the observation set since visual dissimilarity of observations containing the same underlying state is hard to reconcile with reconstruction-based losses.  
 We employ and compare  different unsupervised and  prior-based, i.e., using the similarity signal $s$, models as follows. %listed in the following. 
 
\noindent \emph{i)} The classic \textit{Principal Component Analysis (PCA)} method~\cite{hotelling1933analysis} is used as an unsupervised baseline  method. It obtains the latent mapping by finding the eigenvectors with the highest eigenvalue from the dataset covariance matrix. % ($XX^T$). 
\\\emph{ii)} \textit{Auto-Encoder (AE)}~\cite{kramer1991nonlinear} is considered as another unsupervised approach. AE is  composed  of an encoder and a decoder network  trained jointly
to minimize the Mean Squared Error (MSE) between the input $o$ and decoded output~$\tilde{o}$: $$\mathcal{L}_{ae}(o) \!=\! (o-\tilde{o})^2.$$ % {\ml this loss can be inline}: 
% \begin{equation*}
%     \mathcal{L}_{ae}(o) \!=\! (o-\tilde{o})^2. 
% \end{equation*}
%
\noindent\emph{iii)} A standard \textit{$\beta$-Variational Auto-Encoder (VAE)}~\cite{higgins2016beta} 
is considered as an additional unsupervised model. Similarly to the AE, the  $\beta$-VAE consists of an encoder and a decoder network which are jointly trained  to embody the approximate posterior distribution $q(z|o)$ and the likelihood function $p(o|z)$  providing generative capabilities. The following loss function is minimized:  %by minimizing the following loss function:
\begin{equation*}
        %\resizebox{1\hsize}{!}{$
        \mathcal{L}_{\beta-vae}(o) \!=\! E_{z \sim q(z|o)}[\log p(o|z)] + \beta \!\cdot\! D_{KL}\left(q(z|o) || p(z)\right)
        %$} 
        \label{eq:vaeloss}
    \end{equation*}
 with $z$ the latent variable, $p(z)$ the prior distribution  realized as a standard normal distribution and $D_{KL}(\cdot)$ the KL divergence. 
\\ \noindent \emph{iv)} The similarity signal
 can be exploited through a Pairwise Contrastive (PC) loss~\cite{lsriros}, encouraging the encodings of action pairs  to be larger than a certain distance while minimizing the distance between  similar 
 %no-action 
 pairs. This loss is used to augment the standard AE loss as follows \cite{goroshin2015unsupervised}:  
\begin{equation*}
        %\resizebox{1\hsize}{!}{$
        \mathcal{L}_{pc-ae}(o_i,o_j, s) \!=\! \frac{1}{2}\left(\mathcal{L}_{ae}(o_i)+\mathcal{L}_{ae}(o_j)\right) + 
        \alpha \mathcal{L}_{pc}(o_i, o_j, s)
        %$ } 
        \label{eq:ae_action_loss}
    \end{equation*}
with $\alpha$ a hyperparameter and  $\mathcal{L}_{pc}(o_i, o_j,s) $ %the pairwise contrastive loss 
defined as
\begin{equation}\label{eq:acloss}
%\resizebox{.9\hsize}{!}{$
\mathcal{L}_{pc}(o_i, o_j, s) \!=\! \begin{cases}
			\max(0, d_m - ||z_i-z_j||_1^2) & \text{if } s = 0\\
            ||z_i - z_j||_1^2 & \text{if } s = 1
		 \end{cases}
		 %$}
\end{equation}
where $d_m$ is a hyperparameter denoting the minimum distance that is encouraged between encodings of the  action pairs. We denote the resulting model as \textit{PC-AE}. % and its loss thus combines the reconstruction loss of the standard AE with the pairwise contrastive loss.
%\subsection{Pairwise Contrastive Variational Auto-Encoder  (PC-VAE)}
% The standard $\beta$-VAE is augmented with  an additional pairwise contrastive loss~\cite{lsriros}, called action loss, encouraging the encodings of action pairs at a certain distance while minimizing the distance between no-action pairs. The resulting overall loss is:
%
\\ \noindent \emph{v)} Similarly to the PC-AE, the  task priors
 %weak supervision 
 can also be used to combine the $\beta$-VAE loss with the PC loss, leading to the  following loss function~\cite{lsriros}
\begin{equation*}
%\resizebox{0.99\hsize}{!}{
%$
\begin{aligned}
        \mathcal{L}_{pc-vae}(o_i,o_j, s) =& \frac{1}{2}\left(\mathcal{L}_{\beta-vae}(o_i)
        + \mathcal{L}_{\beta-vae}(o_j) \right) \\ &+
        \gamma \mathcal{L}_{pc}(o_i, o_j, s)  \label{eq:vae_action_loss}
\end{aligned}
%$}
    \end{equation*}
%    {\red 2 lines?}
with $\gamma$ a hyperparameter. We denote this model as \textit{PC-VAE}. 
% The \textit{PC-VAE} therefore combines the reconstruction, KL, as well as the pairwise contrastive loss into a single model.
\\ \noindent  \emph{vi)} A pure contrastive-based model is then considered which is a Siamese network with pairwise contrastive loss~\cite{hadsell2006dimensionality}, referred to as \textit{PC-Siamese}. This model structures the latent space such that it minimizes the pairwise distance between similar pairs and increases it between dissimilar pairs. 
% {\blue Similar pairs are indicated with $s=1$}
%As similar pairs the no-action ($a=0$) pairs are considered,
%while 
As
dissimilar pairs, the action pairs are used ($s=0$). 
This model is based on the sole PC loss $\mathcal{L}_{pc}(o_i, o_j,s) $ in~\eqref{eq:acloss}, i.e., it only relies on the similarity signal 
while no use of reconstruction loss is made.
\\ \noindent  \emph{vii)} A further Siamese network model is considered with different  contrastive loss function. In particular, the following  normalized temperature-scaled Cross Entropy (CE) loss~\cite{chen2020simclear, sohn2016xentloss} is leveraged  which minimizes the cross-entropy between similar pairs using the cosine similarity: 
This model relies on the following normalized temperature-scaled cross-entropy loss~\cite{chen2020simclear, sohn2016xentloss}:      
%\vspace{-4pt}
\begin{equation} \label{eq:entropy}
%\resizebox{0.9\hsize}{!}{$
\begin{aligned}
         \mathcal{L}_{ce}(o_i, o_j)\! =\! 
         -\log \left(\frac{e^{(\mathrm{sim}(z_i,  z_j)/\tau)}}
         {\textstyle\sum_{k=1}^{2N} \mathds{1}_{k \neq i}e^{(\mathrm{sim}(z_i, z_k)/\tau)}}\right)
\end{aligned}
%$}
 \end{equation}
where $\mathrm{sim}( u ,v) =  u^\top v / \lVert u\rVert \lVert v\rVert$ is 
the cosine similarity, $\mathds{1}$ is the indicator function,  $\tau$  is the 
temperature parameter and $N$ is the number of  similar 
pairs that are given in each batch. The resulting model is denoted by \textit{CE-Siamese}.  We use the training procedure in  \cite{chen2020simclear} where, for every  similar
%no-action 
pair, the rest $2(N-1)$ examples are considered dissimilar  as in~\eqref{eq:entropy}.  %{\red Note that this model requires only  no-action pairs $(a=0)$ to be given since dissimilar pairs are implicitly generated. }

 %\begin{wraptable}{rt}{0.45\textwidth}
 %\vspace{-\baselineskip}
\begin{table}[t]
\centering
\resizebox{0.9\linewidth}{!}{
\begin{tabular}{l|ccc}
\hline
Model      & Recon. loss                  & KL loss                   & Contr. loss               \\ \hline
PCA        &   &   &   \\ \hline
AE         & \cmark &   &   \\ \hline
$\beta$-VAE & \cmark & \cmark &   \\ \hline
PC-AE      & \cmark &   & \cmark \\ \hline
PC-VAE     & \cmark & \cmark & \cmark \\ \hline
PC-Siamese   &   &   & \cmark \\ \hline
CE-Siamese   &   &   & \cmark \\ \hline
\end{tabular}
}
%\vspace{-2pt}
\caption{Summary of the considered models with respect to their loss functions.   }
\label{tab:models}

%\vspace{-\baselineskip}
\end{table}
%\end{wraptable}
%\subsection{Models Summary}
\noindent
{\bf Models Summary:}
As summarized in \autoref{tab:models}, the considered models allows to cover a wide range of losses. % which rely on different aspects. % that allows us to gain meaningful insights.
The PCA model is employed as a simple baseline to show that the tasks at hand have adequate complexity and cannot be  solved with a PCA model. The AE and $\beta$-VAE models are mostly based  on the reconstruction loss and therefore  implicitly assume that a visible change in the observations corresponds to a state change. The latter models are then augmented in the PC-AE and PC-VAE models with a pairwise contrastive loss  which exploits the  task priors   ameliorating the visual similarity assumption.  
In addition, PC-Siamese and CE-Siamese  only rely on a contrastive loss without generative capabilities. However, the latter are often not required for downstream tasks.  For the sake of completeness,  in our experiments in \autoref{sec:exp}, we also compare to the case in which no model is used, and raw observations are directly exploited.
    
\section{Latent Planning}\label{sec:planning}
%{\red ML: we can expand this a bit now (rev 2 asks for pseudocode) or include it in Planning Evaluation section? MW: I would rather add the box task. \\}
 As we are interested in ultimately use learned  representations for task planning, we leverage planning in the latent space as a quality measure  itself of the representation, 
 as detailed in the following section. 
 %To this aim, 
 We resort to our latent space planning method from~\cite{lsriros} that builds a graph structure in the latent space, called Latent Space Roadmap (LSR).
  Algorithm~\ref{alg::fix_epsilon} shows a  high level description of the LSR building procedure. 
 \begin{algorithm} \caption{Adapted LSR building~\cite{lsriros}}
 \small
 \begin{algorithmic}[section]
 
 \Require Dataset $\mathcal{T}_z$, min cluster size $m$
 \State $\mathcal{G} =$ build-reference-graph($\mathcal{T}_z$)  \hspace{1.025cm}\# Phase 1
 \State $\mathcal{C}_{z} =$ HDBSCAN-clustering($\mathcal{T}_z$, $m$) \hspace{0.2cm} \# Phase 2
 \State $ \text{LSR} = $ build-LSR($\mathcal{G}$, $\mathcal{C}_{z}$) \hspace{1.55cm} \# Phase 3 \\
 \Return LSR
 \end{algorithmic}
 \label{alg::fix_epsilon}
 \end{algorithm}
 
 The basic idea is to  first build a reference graph using the encodings of action and similar 
 %no-action 
 pairs in $\mathcal{T}_z$ (Phase 1), i.e., nodes are created for each encoding  and they are connected in case of action pairs. Next, in Phase 2, the latent space is clustered. We substitute the $\varepsilon$-clustering used in~\cite{lsriros} with the HDBSCAN~\cite{mcinnes2017hdbscan} which only requires the minimum samples  per cluster $m$ to be set. %for phase 2 instead the $\varepsilon$-clustering used in~\cite{lsriros}. 
 The LSR is  then built in Phase~3 where each cluster is represented by a node that embodies the underlying state and clusters are connected through edges if they  %cluster the latent space, associate a node to each cluster and build edges between the nodes if they 
 are one action apart, i.e., they contain encodings of action pairs.
 To use the LSR for planning, we first encode the start and goal observations with the model of interest and then select the respective closest nodes in the LSR as start and goal nodes of the path. Finally, we find the shortest paths from the start node to the goal one.  %{\red In this work, we use the HDBSCAN~\cite{mcinnes2017hdbscan} clustering method for the LSR which only requires the minimum samples per cluster to be set.} 
  Note that the objective of the planning is to produce a sequence of \emph{actions} %(plan) 
  that lead from start to goal states. No decoded images are then needed and the LSR can be built in the latent space generated by any model in \autoref{sec:models}. 

 \section{Representation Evaluation Metrics}\label{sec:scores}
To evaluate the performance of the different latent mapping models, we propose two types of metrics.
First, as stated in \autoref{sec:problem}, the structure of the latent space can be assessed through \emph{clustering}, i.e., a good latent space should be easy to cluster. Second, the latent space should be  suitable for task planning - a good latent space should result in easier planning. Thus, we also evaluate the \emph{planning} performance of learned representations.  
%\vspace{-\baselineskip}
\subsection{Clustering metrics}

\noindent
\textbf{Homogeneity \& Completeness:} Given the ground truth states, the homogeneity score \cite{rosenberg2007v}, denoted by $h_c$,  measures the purity of the created clusters, i.e.,  that each cluster contains elements from the same state. %On the other hand, 
Completeness, denoted by $c_c$, measures the preservation of the states in clusters, i.e., that each state is assigned to the same cluster. Both the metrics have range  $[0,1]$,  with $1$ being the best value. Assigning all elements in different clusters would results in $h_c = 1$ and $c_c = 0$, while assigning all elements in the same cluster would results in $h_c = 0$ and $c_c = 1$. These quantities are calculated based on cross-entropy as formulated in~\cite{rosenberg2007v}.

\noindent
\textbf{Mean silhouette coefficient:} The silhouette coefficient \cite{rousseeuw1987silhouettes}, denoted by $s_c^i$,  is defined for each sample $i$ and, in contrast to the previous metrics,  does not rely on ground truth labels. Let $d_{intra}^i$ be the  mean distance between  sample $i$ and all the other points in the same  cluster  and let $d_{closest}^i$ be the mean distance between  sample $i$ and all other points in the closest cluster. The silhouette coefficient for sample $i$ is defined as:
 $$s_c^i= \frac{(d_{closest}^i-d_{intra}^i)}{ \max(d_{intra}^i,d_{closest}^i)}$$
%  $$s_c^i= (d_{closest}^i-d_{intra}^i) / \max(d_{intra}^i,d_{closest}^i)$$
which can assume values in $[-1,1]$, with higher values indicating dense and well-separated clusters. We report the mean silhouette coefficient $s_c$ over all samples.  

%\vspace{-\baselineskip}
\subsection{Planning Evaluation}

To assess the planning performance achieved through the LSR, we evaluate both 
%\st{the LSR edges quality}
graph structure
and the obtained start to goal paths.
We define the \emph{true} representative state for each node in the LSR  as the state that is contained the most. The following metrics are considered:

\noindent
\textbf{Number of Nodes}:  It is the number of nodes in the LSR and is denoted by $|\mathcal{V}|$. This number should ideally be equal to the number of possible underlying states of the system.

 \noindent
\textbf{Number of Edges}:  It represents the number of edges that are built between nodes in the LSR and is denoted by $|\mathcal{E}|$. % according to the method in Sec.~\ref{sec:planning}.   
In the case of optimal latent mapping and graph, the number of edges should be equal to the number of possible transitions between states of the system.

\noindent
 \textbf{Correctness of the Edges}: It is denoted by $c_e$ and quantifies how many nodes are improperly connected in the LSR. In detail, it is defined as the number of \emph{legal} edges, i.e., the edges associated to allowed state transitions according to the task rules, divided by the total number of edges. This score has range $[0,1]$, with $1$ being the best value.

\noindent
\textbf{Path Metrics}:   To evaluate the latent planning capabilities, we evaluate the correctness of the shortest paths between random novel start and goal observations (taken from holdout datasets). We consider $1000$ different start and goal observations and evaluate the percentage that all found paths are correct, denoted by \textit{\% all},  and the percentage that at least one path is correct, denoted by \textit{\% any}.

 %\vspace{-\baselineskip}
 \section{Validation Setting }\label{sec:val-setting}
 \begin{figure}[t]
  \centering
  \includegraphics[width=0.95\linewidth]{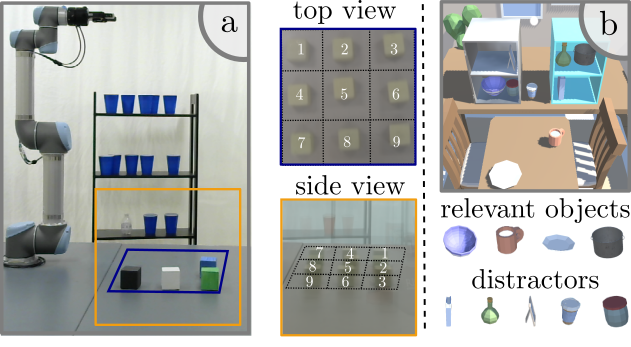}
  \caption{\textbf{a)} Box manipulation dataset with two 
  viewpoints. \textbf{b)}  % along with three viewpoints. Wall and table colors are considered irrelevant for the task. 
  Shelf arrangement dataset with the task relevant objects (top object row) and the five distractor objects (bottom object row).
  }\label{fig:datasets}
%   \caption{ Left: Box stacking dataset  with three different viewpoints. Right:  % along with three viewpoints. Wall and table colors are considered irrelevant for the task. 
%   Shelf arrangement dataset with the task relevant objects (top object row) and the five distractor objects (bottom 
%   object row).
%   }
%\vspace{-\baselineskip}
\end{figure}
%\end{wrapfigure}

 %Two tasks are considered in a simulated environment: \emph{i)}~a box stacking task and a \emph{ii)}~shelf arrangement task. 
 % {\blue 
 Two tasks are considered: a box manipulation task on a real robotic system, and a simulated shelf arrangement, %and a simulated box stacking task 
 in Unity~\cite{unitygameengine} environment.  An additional simulated box stacking task can be found in our preliminary workshop paper \cite{chamzas2020state} as well as in 
 Appendix~\ref{app:sim_box_stacking_task}. 
  It is worth highlighting that the goal of this work is not to solve these tasks in an innovative manner, but rather to gain general insights that can be transferred to cases where a determination of the exact underlying state is not possible.
 
  In  each task, the \emph{task-relevant} objects are interchangeable -- i.e., swapping two objects results in the same state. 
 Their arrangement in the scene specifies the underlying state of the system. Other objects that are irrelevant for the task, referred to as \emph{distractor} objects, can be present in the observations.
 The objective of all tasks is to plan a sequence of states to arrange the \emph{relevant} objects according to a goal observation. Transitions between states -- i.e., actions, can be then retrieved through the LSR~\cite{lsriros}. 
 All datasets are available on the website\footref{fn:website}. 
 
\begin{table*}[t!]
\resulttable{ $\mathcal{BM}_{t}$}{
- & $1016$ & $0.92$ & $0.92$ & $0.79$ & $583$ & $0.78$ & $0.0$ & $0.0$
\\ \hline 
PCA &  $496$ & $0.75$ & $0.78$ & $0.52$ & $452$ & $0.52$ & $0.0$ & $0.0$
\\ \hline 
AE &  $233$ & $0.49$ & $0.57$ & $0.29$ & $234$ & $0.27$ & $0.0$ & $0.0$
\\ \hline 
$ \beta$-VAE  &  $539$ & $0.85$ & $0.85$ & $0.51$ & \boldmath$422$ & $0.62$ & $0.0$ & $0.0$
\\ \hline 
PC-AE &  $246$ & $0.54$ & $0.6$ & $0.3$ & $258$ & $0.28$ & $0.0$ & $0.0$
\\ \hline 
PC-VAE &  $570$ & $1.0$ & \boldmath$1.0$ & $0.58$ & $488$ & \boldmath$1.0$ & $29.9$ & $29.9$
\\ \hline 
PC-Sia. &  $389$ & $0.99$ & \boldmath$1.0$ & $0.52$ & $458$ & $0.97$ & $47.37$ & $57.3$
\\ \hline 
CE-Sia. &  \boldmath$150$ & \boldmath$1.0$ & \boldmath$1.0$ & \boldmath$0.67$ & $325$ & \boldmath$1.0$ & \boldmath$98.3$ & \boldmath$98.3$
\\ \hline 
 \hline
&  \multicolumn{8}{c|}{Dataset $\overline{\mathcal{BM}}_{t}$}\\ \hline
PC-AE &  $218$ & $0.99$ & \boldmath$1.0$ & $0.71$ & $375$ & $0.98$ & $72.12$ & $82.5$
\\ \hline 
PC-VAE &  $395$ & \boldmath$1.0$ & \boldmath$1.0$ & $0.56$ & \boldmath$461$ & \boldmath$1.0$ & $89.7$ & $89.7$
\\ \hline 
PC-Sia. &  \boldmath$133$ & \boldmath$1.0$ &\boldmath $1.0$ & \boldmath$0.9$ & $314$ & \boldmath$1.0$ & \boldmath$97.7$ & \boldmath$98.2$
\\ \hline
}
\resulttable{$\mathcal{BM}_{st}$}{
- & $710$ & $0.83$ & $0.83$ & $0.5$ & $496$ & $0.58$ & $0.0$ & $0.0$ 
\\ \hline 
 PCA &  $400$ & $0.59$ & $0.62$ & $0.46$ & \boldmath$453$ & $0.25$ & $0.0$ & $0.0$
\\ \hline 
AE &  $554$ & $0.87$ & $0.88$ & $0.56$ & $454$ & $0.69$ & $0.0$ & $0.0$
\\ \hline 
$ \beta$-VAE  &  $407$ & $0.72$ & $0.74$ & $0.44$ & $361$ & $0.38$ & $0.0$ & $0.0$
\\ \hline 
PC-AE &  $318$ & $0.62$ & $0.91$ & \boldmath$0.61$ & $325$ & $0.47$ & $0.0$ & $0.0$
\\ \hline 
PC-VAE &  $381$ & $0.84$ & $0.85$ & $0.42$ & $295$ & $0.65$ & $0.1$ & $0.1$
\\ \hline 
PC-Sia. &  $289$ & $0.96$ & $0.96$ & $0.4$ & $312$ & $0.92$ & $26.34$ & $27.5$
\\ \hline 
CE-Sia. &  \boldmath$232$ & \boldmath$0.99$ & \boldmath$0.99$ & $0.41$ & $354$ & \boldmath$0.99$ & \boldmath$78.39$ & \boldmath$78.7$
\\ \hline 
 \hline
&  \multicolumn{8}{c|}{Dataset  $\overline{\mathcal{BM}}_{st}$} 
\\ \hline
PC-AE &  $158$ & $0.93$ & $0.98$ & \boldmath$0.5$ & \boldmath$310$ & $0.79$ & $26.01$ & $36.4$
\\ \hline 
PC-VAE &  $164$ & $0.89$ & $0.94$ & $0.36$ & $198$ & $0.77$ & $7.39$ & $9.2$
\\ \hline 
PC-Sia. &  \boldmath$136$ & \boldmath$0.99$ & \boldmath$0.99$ & $0.45$ & $282$ & \boldmath$0.98$ & \boldmath$69.37$ & \boldmath$72.4$
\\ \hline 
}
\caption{
 Evaluation results for the latent mapping models  and raw observations  on  $\mathcal{BM}_{t}$  and $\mathcal{BM}_{st}$ (top) and their augmented versions (bottom) $\overline{\mathcal{BM}}_{t}$  and $\overline{\mathcal{BM}}_{st}$ for the box manipulation task. Best results in bold. }
\label{tab:results_box_pushing}
%\vspace{-\baselineskip}
\end{table*}
 
\noindent
\textbf{Box Manipulation:} 
The setup of this real-world case study is shown in \autoref{fig:datasets}a). The task is composed of four interchangeable boxes, and each box can only move to adjacent tiles in   a $3 \times 3$ grid. The robot is tasked with moving the boxes to the state of the goal image. 
This task has $126$ possible states with $420$ allowed state transitions.
Two different viewpoints are considered to capture the scene and three datasets are built as follows: \emph{i)} $\mathcal{BM}_s$, where all the observations are taken from the side view (in orange in \autoref{fig:datasets}), \emph{ii) } $\mathcal{BM}_t$, where the observations are only taken from the top view (in blue in \autoref{fig:datasets}), and \emph{iii)} $\mathcal{BM}_{st}$, where views are randomly picked from the side or top view. 
Images have naturally occurring task-irrelevant factors such as distractor objects changing in the background (side view), as well as out-of-focus images. In \autoref{fig:datasets}a) the mean image of all training images for side and top view are depicted.  In the following, we report the considered self-supervised data collection procedure.  

 \textit{Real world Training}.  As actions, we employ pick and place  operations realized by the following sequence: moving the robot, through a motion planner, to the pick location, closing the gripper, moving to the place location,  and opening the gripper. To generate an action pair, the robot performs a random action  -- moves a box to an adjacent tile. To create similar states, it swaps two boxes.  The swapping is simply three consecutive pick and place operations. Before executing each action the robot needs to check that the preconditions of that action are true, e.g., pick location is occupied and place location is empty. This can be verified by moving and closing the gripper to the pick and place locations.
If the gripper fully closes (sensed through the gripper encoder), the location is empty,  otherwise a box is present and can be picked. A similar verification could be achieved with a depth camera. This formulation is consistent with the high-level actions in~\cite{konidaris2018skills}.
Using this procedure $2800$ training data samples with $1330$ action pairs were collected in a self-supervised manner by randomly performing actions. Note that no access to the underlying state nor human labeling is required to generate this dataset. 
See the supplementary video for more details. % {\red on this task. - can be killed}
%Appendix \ref{sec:app:exp_box_push} and 

\noindent
\textbf{Shelf Arrangement:} As depicted in \autoref{fig:datasets}b), the scene  of the shelf arrangement task is composed of two shelving units with a total of four shelves,
and a table where four objects can be placed. Four \emph{task-relevant} objects -- a bowl, a pot, a mug and a plate (shown in the figure) -- are present in the scene. 
This task has $70$ possible states and $320$ possible transitions. In addition, \emph{distractor} objects (bottom right part of Fig.~\ref{fig:datasets})  can be present on the shelves and change their position. 
 Two datasets are thus defined: \emph{i)} $\mathcal{SA}_{0d}$, that contains the four relevant objects and zero distractor objects  ($2500$ tuples with $1240$ action pairs), 
\emph{ii)} $\mathcal{SA}_{5d}$, that contains all five distractor objects with each distractor having a probability of $0.8$ to appear on the shelf ($2500$ tuples with $1277$ action pairs).
{ For more information about the tasks and their rules see Appendix~\ref{app:pushing_rules}-\ref{app:stacking_rules_shelf}.}

We trained each of the seven models in Sec.~\ref{sec:models} (PCA, AE, $\beta$-VAE, PC-AE, PC-VAE, PC-Siamese, and CE-Siamese) with the datasets of the above defined tasks 
($\mathcal{BM}_s$, $\mathcal{BM}_t$, $\mathcal{BM}_{st}$ for box manipulation; $\mathcal{SA}_{0d}$, $\mathcal{SA}_{5d}$ for shelf arrangement) as well as their augmented versions  ($\overline{\mathcal{BM}}_s$, $\overline{\mathcal{BM}}_t$, $\overline{\mathcal{BM}}_{st}$, $\overline{\mathcal{SA}}_{0d}$, $\overline{\mathcal{SA}}_{5d}$),    with $n=1$ in Sec.~\ref{sec:problem}. 
The evaluation was performed on respective holdout datasets composed of $334$ and $2500$ novel tuples, respectively.
Further details on the architectures,  hyperparameters,  and additional plots can be found { 
in the Appendix~\ref{app:architecture},}
the  website\footref{fn:website}, and the code\footnote{\label{fn:code}\mbox{\url{https://github.com/State-Representation/code}}}. %{\red ML: should we add some implementation details? MW: I think refrencing the code and website is enough. }

\section{Results and Discussion}\label{sec:exp}
Two main questions are discussed in detail:
\begin{compactenum}
    \item Do contrastive-based losses outperform reconstruction-based losses  when task-irrelevant factors of variations are present in the observations?  
    \item Can simple data augmentation as described in  Sec.~\ref{sec:problem} 
    boost the representation performance?  
\end{compactenum}

\begin{figure*}[t]
  \centering
  \includegraphics[width=0.98\textwidth]{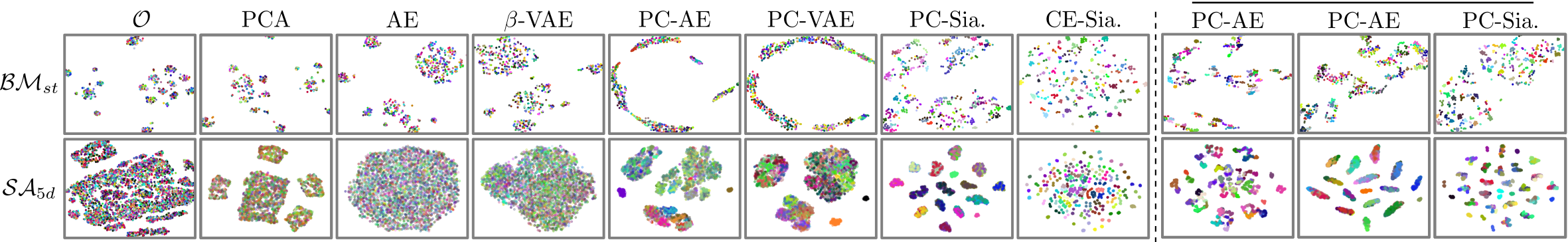}
  %\vspace{-5pt}
  \caption{ Two-dimensional t-SNE plots for the box manipulation task for the mixed view (top row) and the shelf arrangement task for the distractor case (bottom row). Each color is associated with a possible underlying state. On the right we display the plots for the augmented contrastive models.  Full results are  accessible on
  %in the App.~\ref{sec:app:exp_box_push}-\ref{sec:app:exp_shelf} and} 
  the website\footref{fn:website}. 
}
  \label{fig:t-SNEs_results_all}
 % \vspace{-\baselineskip}
\end{figure*}

\begin{table*}[htb!]
\resulttable{ $\mathcal{SA}_{0d}$}{
PC-AE &  $5$ & $0.28$ & \boldmath$1.00$ & $0.93$ & $4$ & $0.75$ & $0.00$ & $0.00$
\\ \hline 
PC-VAE &  $5$ & $0.28$ & \boldmath$1.00$ & $0.78$ & $4$ & $0.75$ & $0.00$ & $0.00$
\\ \hline 
PC-Sia. &  \boldmath$16$ & $0.64$ & \boldmath$1.00$ & \boldmath$0.96$ & \boldmath$32$ & $0.59$ & $1.01$ & $1.8$
\\ \hline 
CE-Sia. &  $296$ & \boldmath$1.00$ & \boldmath$1.00$ & $0.54$ & $842$ & \boldmath$1.00$ & \boldmath$95.90$ & \boldmath$95.90$
\\ \hline 
 \hline
&  \multicolumn{8}{c|}{Dataset $\overline{\mathcal{SA}}_{0d}$} \\ \hline
PC-AE &  $97$ & $0.87$ & $0.95$ & $0.51$ & \boldmath$368$ & $0.77$ & $20.30$ & $35.90$
\\ \hline 
PC-VAE &  \boldmath$64$ & $0.90$ & $0.99$ & $0.32$ & $235$ & $0.77$ & $33.13$ & $55.40$
\\ \hline 
PC-Sia. &  $225$ & \boldmath$1.00$ & \boldmath$1.00$ & \boldmath$0.74$ & $772$ & \boldmath$1.00$ & \boldmath$100.0$ & \boldmath$100.0$
\\ \hline 
}
\resulttable{$\mathcal{SA}_{5d}$}{
PC-AE &  $5$ & $0.28$ & \boldmath$1.00$ & $0.9$ & $4$ & $0.75$ & $0.40$ & $0.40$
\\ \hline 
PC-VAE &  $5$ & $0.28$ & \boldmath$1.00$ & $0.79$ & $4$ & $0.75$ & $0.40$ & $0.40$
\\ \hline 
PC-Sia. &  \boldmath$16$ & $0.64$ & \boldmath$1.00$ & \boldmath$0.98$ & \boldmath$32$ & $0.69$ & $2.42$ & $4.10$
\\ \hline 
CE-Sia. &  $286$ & \boldmath$1.00$ & \boldmath$1.00$ & $0.46$ & $841$ & \boldmath$0.99$ & \boldmath$95.12$ & \boldmath$95.30$
\\ \hline  
\hline
&  \multicolumn{8}{c|}{Dataset  $\overline{\mathcal{SA}}_{5d}$} 
\\ \hline
PC-AE &  $18$ & $0.49$ & $0.98$ & $0.01$ & $34$ & $0.44$ & $0.24$ & $0.40$
\\ \hline 
PC-VAE &  $16$ & $0.64$ & \boldmath$1.00$ & $0.52$ & $32$ & $0.69$ & $2.42$ & $4.10$
\\ \hline 
PC-Sia. &  \boldmath$30$ & \boldmath$0.78$ & \boldmath$1.00$ & \boldmath$0.61$ & \boldmath$87$ & \boldmath$0.74$ & \boldmath$6.66$ & \boldmath$13.10$
\\ \hline 
}
%\vspace{-1em }
%\vspace{-5pt}
\caption{
 Evaluation results for the contrastive-based latent mapping models on  $\mathcal{SA}_{0d}$ (top left) and $\mathcal{SA}_{5d}$ (top right) and their augmented versions $\overline{\mathcal{SA}}_{0d}$ (bottom left) and $\overline{\mathcal{SA}}_{5d}$ (bottom right) for the shelf arrangement task. Best results in bold. }
\label{tab:results_shelf_arranging}
%\vspace{-\baselineskip}
%\vspace{-1pt }
\end{table*}

 \noindent
 \textbf{Influence of Contrastive Loss:}
 To answer question $1$, 
we carry out a quantitative and a qualitative analysis on the box manipulation task. 
The former is summarized in  Table~\ref{tab:results_box_pushing} (top part). 
 We observe that models PC-VAE, PC-Siamese and CE-Siamese, employing a contrastive loss, manage to achieve almost perfect performance in terms of homogeneity ($h_c$),  completeness ($c_c$) and edge score ($c_e$) with top view dataset $\mathcal{BM}_t$, enabling planning in their latent spaces. In particular, best planning performance ($98.3\%$ for $\% \,any$) is achieved by the pure contrastive model CE-Siamese, followed by PC-Siamese ($57.3\%$ for $\% \,any$) and PC-VAE ($29.9\%$ for $\% \,any$). 
  In contrast, the case of no latent mapping (first row), i.e., the use of raw observations,  the PCA and the reconstruction-based models achieve very low clustering and planning performance, reaching no correct paths.  This also confirms the unsuitability of directly using  raw high-dimensional observations for task planning with task-irrelevant factors of variations.  Similarly,  model PC-AE obtains poor performance, reporting $c_e=0.28$ which leads the planning to fail due to an excessive number of erroneous edges. This suggests that the sole addition of the contrastive loss to the reconstruction one may be not sufficient to effectively structure the latent space.  
For the dataset $\mathcal{BM}_{st}$ (top right), having  observations taken from different viewpoints, it can be noticed that the pure contrastive-based models CE-Siamese and PC-Siamese obtain the best performance in terms of clustering and planning, with CE-Siamese ($78.7\%$ for $\% \,any$) outperforming PC-Siamese  ($27.5\%$ for $\% \,any$), 
while zero success correct paths are obtained by PC-VAE and PC-AE, mixing reconstruction and contrastive losses, as well as by PCA, AE and $\beta$-VAE. 
This confirms the relevance of leveraging task priors %weak supervision 
to handle task-irrelevant factors of variation, like the different viewpoints. 
The effectiveness of the best performing model (CE-Siamese) in regards to planning was also validated on the real robotic system shown in the supplementary video.
{ Results with  $\mathcal{BM}_s$ can be found in Appendix~\ref{sec:app:exp_box_push}. }

The above results are also reflected in the qualitative analysis in \autoref{fig:t-SNEs_results_all} (top row, left part) where 
  the  latent encodings obtained with the different models  as well as raw observations ($\mathcal{O}$ column) are 
  visualized through $2$D t-SNE~\cite{maaten2008visualizing} plots. 
 Results with $\mathcal{BM}_{st}$ are shown. 
We can notice that  the raw observations, PCA and purely reconstruction-based models AE and $\beta$-VAE fail in structuring the representations, forming spurious clusters in which different states are mixed up. Non-homogeneous clusters are also obtained by PC-AE and PC-VAE, while a significant improvement of the latent space structure is recorded by the purely contrastive loss based Siamese networks (PC-Siamese and CE-Siamese), leading to visually distinct clusters. 
\\\textit{In summary,  we observe that the contrastive-based models  (PC-Siamese, CE-Siamese) 
outperform the other ones by a significant margin. Notably, the architectures of the Siamese networks are much shallower\footref{fn:code} than the AE and VAE ones,  leading to considerably faster training processes} ( $< 3.5$ minutes vs  $\approx 2.5$ hours on a  GeForce GTX $1080$ Ti).

 %\subsection{Influence of Data Augmentation} % on Latent Space Structure}
%\subsection{Effect of Data Augmentation on Latent Representations}
 \noindent
 \textbf{Influence of Data Augmentation:}
 To evaluate the influence of the data augmentation in   Sec.~\ref{sec:problem}, we first analyze the representation performance on the 
 shelf arrangement task when it is applied and when it is not.
 For the sake of space, we focus only on the four contrastive-based models since the unsuitaibility of raw observations, PCA, AE and $\beta$-VAE has been shown above. 
% very poor performance is achieved by {\red using raw observations,} PCA, AE and $\beta$-VAE. {\red as their effectiveness has been shown above.? or the unsuitaibility of ... }
% {\blue For more plots and results visit the project website. {\red ML: I think this should be killed}}
 { Full results can be found in Appendix~\ref{sec:app:exp_shelf}.}
 Table~\ref{tab:results_shelf_arranging} reports the obtained evaluation metrics.  
When no augmentation is applied (top part), all the models, except for  CE-Siamese, show very low performance  for both clustering and planning on both datasets, creating a small number of clusters ($\ll 70$) that are erroneously connected. In contrast, CE-Siamese generates a large amount of pure clusters ($ \simeq  300$ clusters with $h_e  \simeq  1$ for both datasets) which are almost perfectly connected ($c_e  \simeq  1$), leading to high path metrics (\textit{\% any} $ \simeq  95\%$ for both datasets). 
%{\blue The general low performance in the shelf arrangement task motivates why we tested the augmentation on it. }
% We thus focus on the shelf arrangement task to evaluate the effect of the augmentation technique as a greater performance improvement is possible compared to the box stacking task. 
When the augmentation is used, the performance of all models improves 
 for the no distractors dataset (bottom left), leading the PC-Siamese to reach perfect path metrics ($100\%$ for \textit{\%~any}) and PC-AE and PC-VAE to reach $ \simeq  36\%$ and $ \simeq  55\%$ for  \textit{\% any}. This confirms the beneficial effect of the considered augmentation which, however,
is not equivalently effective when distractor objects are present in the scene (bottom right).   More specifically, only PC-Siamese is positively influenced by the augmentation with $\overline{\mathcal{SA}}_{5d}$, reaching path metrics \textit{\% any} $ \simeq  13\%$ (from $ \simeq 4\%$). This suggests that a higher number of dissimilar pairs should be synthetically generated for this case study, i.e., $n\gg 1$.   Note the augmented datasets are only used for the latent mapping but not for the LSR building to avoid building wrong edges.  Moreover,  the CE-Siamese is not evaluated with the augmentation technique since it does not use action pairs.  Similar observations also hold for the box manipulation dataset in Table~\ref{tab:results_box_pushing}, where we can notice that, when the augmentation is used (bottom part), PC-Siamese manages to achieve almost perfect performance on the top view  dataset $\mathcal{BM}_t$, with $|\mathcal{V}|=133$ clusters and path performance $98.2\%$ for $\% \,any$, as well as good performance on the mixed view dataset $\mathcal{BM}_{st}$, with  path performance $78.8\%$ for $\% \,any$. General improvements are also recorded for PC-AE and PC-VAE which, however, underperform the purely contrastive-based models.    

\autoref{fig:t-SNEs_results_all} reports the t-SNE plots for the shelf stacking task (bottom row) with five distractors ($\mathcal{SA}_{5d}$) obtained with (on the right) and without (on the left) augmentation.
In this task the optimal number of clusters is $70$. 
 It is evident from the t-SNE visualizations that, in the absence of augmentation, only the CE-Siamese model can structure the encodings such that clusters of different states are not overlapping.   
This is due to the training procedure of CE-Siamese~\cite{chen2020simclear}, which only relies on  similar
%no-action 
pairs and synthetically builds a large number of dissimilar pairs \cite{chen2020simclear}. In contrast,  better separation of the states is observed with data augmentation. 
Notably, in $\overline{\mathcal{SA}}_{5d}$,  PC-Siamese, which solely relies on the contrastive loss, achieves a better clustering than PC-AE and PC-VAE, which also exploit reconstruction loss. Similar considerations also hold for the box manipulation task (top row of \autoref{fig:t-SNEs_results_all}). 
\textit{In summary, we observe that a simple data augmentation boosts the performance of the contrastive-based models. }
\section{Conclusion}
 In this work, we investigated the effect of different loss functions for retrieving the underlying states of a system from visual observations  applied to task planning.
 We showed that purely reconstruction-based models  are prone to fail when task-irrelevant factors of variation are present in the observations.
 In contrast, the exploitation of  task priors 
 %{\red weak supervision} 
 in  contrastive-based losses as well as of an easy data augmentation technique resulted in a significant representation improvement. We analyzed two  robotics tasks with different task-irrelevant factors of variation: \emph{i)} box manipulation, on a real robotic system with different viewpoints and occlusions, and
% \emph{ii)} simulated box stacking {\red (shown in the Appendix~\ref{app:sim_box_stacking_task})}, with different viewpoints and colors of background and table, and 
 \emph{ii)} shelf arrangement, with distractor objects that  are irrelevant for the task itself.  
% {\blue The results confirm  our initial insights gained when analysing a simulated box stacking task with different viewpoints and colors of background and table in~\cite{chamzas2020state}, with now a large scale evaluation on more relevant and complex tasks.}
 We thus believe that contrastive-based losses as well as simple data augmentations  go a long way toward obtaining meaningful representations that can be used for a wide variety of robotics tasks  and provide a promising direction for the research community. 

\bibliographystyle{IEEEtran}
\bibliography{references} 

\begin{thebibliography}{10}
\providecommand{\url}[1]{#1}
\csname url@samestyle\endcsname
\providecommand{\newblock}{\relax}
\providecommand{\bibinfo}[2]{#2}
\providecommand{\BIBentrySTDinterwordspacing}{\spaceskip=0pt\relax}
\providecommand{\BIBentryALTinterwordstretchfactor}{4}
\providecommand{\BIBentryALTinterwordspacing}{\spaceskip=\fontdimen2\font plus
\BIBentryALTinterwordstretchfactor\fontdimen3\font minus
  \fontdimen4\font\relax}
\providecommand{\BIBforeignlanguage}[2]{{%
\expandafter\ifx\csname l@#1\endcsname\relax
\typeout{** WARNING: IEEEtran.bst: No hyphenation pattern has been}%
\typeout{** loaded for the language `#1'. Using the pattern for}%
\typeout{** the default language instead.}%
\else
\language=\csname l@#1\endcsname
\fi
#2}}
\providecommand{\BIBdecl}{\relax}
\BIBdecl

\bibitem{konidaris2018skills}
G.~Konidaris, L.~P. Kaelbling, and T.~Lozano-Perez, ``From skills to symbols:
  Learning symbolic representations for abstract high-level planning,''
  \emph{J. Artificial Intelligence Res.}, vol.~61, pp. 215--289, 2018.

\bibitem{stooke2020decoupling}
A.~Stooke, K.~Lee, P.~Abbeel, and M.~Laskin, ``Decoupling representation
  learning from reinforcement learning,'' \emph{arXiv preprint
  arXiv:2009.08319}, 2020.

\bibitem{dantam2016incremental}
N.~T. Dantam, Z.~K. Kingston, S.~Chaudhuri, and L.~E. Kavraki, ``Incremental
  task and motion planning: A constraint-based approach.'' in \emph{Robotics:
  Science and Syst.}, vol.~12, 2016, p. 00052.

\bibitem{thomas2018disentangling}
V.~Thomas, E.~Bengio, W.~Fedus, J.~Pondard, P.~Beaudoin, H.~Larochelle,
  J.~Pineau, D.~Precup, and Y.~Bengio, ``Disentangling the independently
  controllable factors of variation by interacting with the world,''
  \emph{Learning Disentangling Representations Wksp. at NeurIPS}, 2017.

\bibitem{Ichter2019b}
B.~Ichter and M.~Pavone, ``Robot motion planning in learned latent spaces,''
  \emph{{IEEE} Robot. Autom. Letters}, vol.~4, no.~3, pp. 2407--2414, 2019.

\bibitem{watter2015embed}
M.~Watter, J.~Springenberg, J.~Boedecker, and M.~Riedmiller, ``Embed to
  control: A locally linear latent dynamics model for control from raw
  images,'' \emph{Adv. Neural Inf. Process. Syst.}, vol.~28, pp. 2746--2754,
  2015.

\bibitem{hafner2019learning}
D.~Hafner, T.~Lillicrap, I.~Fischer, R.~Villegas, D.~Ha, H.~Lee, and
  J.~Davidson, ``Learning latent dynamics for planning from pixels,'' in
  \emph{Int. Conf. on Mach. Learn.}, 2019, pp. 2555--2565.

\bibitem{pertsch2020long}
K.~Pertsch, O.~Rybkin, F.~Ebert, S.~Zhou, D.~Jayaraman, C.~Finn, and S.~Levine,
  ``Long-horizon visual planning with goal-conditioned hierarchical
  predictors,'' \emph{Adv. Neural Inf. Process. Syst.}, vol.~33, 2020.

\bibitem{hafner2019dream}
D.~Hafner, T.~Lillicrap, J.~Ba, and M.~Norouzi, ``Dream to control: Learning
  behaviors by latent imagination,'' in \emph{Int. Conf. on Learn.
  Representations}, 2019.

\bibitem{nair2020goal}
S.~Nair, S.~Savarese, and C.~Finn, ``Goal-aware prediction: Learning to model
  what matters,'' in \emph{Int. Conf. on Mach. Learn.}, 2020, pp. 7207--7219.

\bibitem{finn2016deep}
C.~Finn, X.~Y. Tan, Y.~Duan, T.~Darrell, S.~Levine, and P.~Abbeel, ``Deep
  spatial autoencoders for visuomotor learning,'' in \emph{{IEEE} Int. Conf.
  Robot. Autom.}, 2016, pp. 512--519.

\bibitem{hoque2020visuospatial}
R.~Hoque, D.~Seita, A.~Balakrishna, A.~Ganapathi, A.~K. Tanwani, N.~Jamali,
  K.~Yamane, S.~Iba, and K.~Goldberg, ``Visuospatial foresight for multi-step,
  multi-task fabric manipulation,'' \emph{Robotics: Science and Syst.}, 2020.

\bibitem{hoof2016stable}
H.~{van Hoof}, N.~{Chen}, M.~{Karl}, P.~{van der Smagt}, and J.~{Peters},
  ``Stable reinforcement learning with autoencoders for tactile and visual
  data,'' in \emph{{IEEE/RSJ} Int. Conf. on Intell. Robots and Syst.}, 2016,
  pp. 3928--3934.

\bibitem{reviewCRL2020}
P.~H. Le-Khac, G.~Healy, and A.~F. Smeaton, ``Contrastive representation
  learning: A framework and review,'' \emph{IEEE Access}, vol.~8, p.
  193907–193934, 2020.

\bibitem{srinivas2020curl}
M.~Laskin, A.~Srinivas, and P.~Abbeel, ``Curl: Contrastive unsupervised
  representations for reinforcement learning,'' in \emph{Int. Conf. on Mach.
  Learn.}, 2020, pp. 5639--5650.

\bibitem{kipf2019contrastive}
T.~Kipf, E.~van~der Pol, and M.~Welling, ``Contrastive learning of structured
  world models,'' in \emph{Int. Conf. on Learn. Representations}, 2019.

\bibitem{tcn2018}
P.~Sermanet, C.~Lynch, Y.~Chebotar, J.~Hsu, E.~Jang, S.~Schaal, S.~Levine, and
  G.~Brain, ``Time-contrastive networks: Self-supervised learning from video,''
  in \emph{{IEEE} Int. Conf. Robot. Autom.}, 2018, pp. 1134--1141.

\bibitem{zhang2020learning}
A.~Zhang, R.~T. McAllister, R.~Calandra, Y.~Gal, and S.~Levine, ``Learning
  invariant representations for reinforcement learning without
  reconstruction,'' in \emph{Int. Conf. on Learn. Representations}, 2020.

\bibitem{jonschkowski2015learning}
R.~Jonschkowski and O.~Brock, ``Learning state representations with robotic
  priors,'' \emph{Autonomous Robots}, vol.~39, no.~3, pp. 407--428, 2015.

\bibitem{wiskott2002slow}
L.~Wiskott and T.~J. Sejnowski, ``Slow feature analysis: Unsupervised learning
  of invariances,'' \emph{Neural computation}, vol.~14, no.~4, pp. 715--770,
  2002.

\bibitem{legenstein2010reinforcement}
R.~Legenstein, N.~Wilbert, and L.~Wiskott, ``Reinforcement learning on slow
  features of high-dimensional input streams,'' \emph{PLoS computational
  biology}, vol.~6, no.~8, p. e1000894, 2010.

\bibitem{hofer2010using}
S.~H{\"o}fer, M.~Hild, and M.~Kubisch, ``Using slow feature analysis to extract
  behavioural manifolds related to humanoid robot postures,'' in \emph{Int.
  Conf. on Epigenetic Robot.}, 2010, pp. 43--50.

\bibitem{lsriros}
M.~Lippi, P.~Poklukar, M.~C. Welle, A.~Varava, H.~Yin, A.~Marino, and
  D.~Kragic, ``Latent space roadmap for visual action planning of deformable
  and rigid object manipulation,'' \emph{{IEEE/RSJ} Int. Conf. on Intell.
  Robots and Syst.}, 2020.

\bibitem{jetchevground2013}
N.~Jetchev, T.~Lang, and M.~Toussaint, ``Learning grounded relational symbols
  from continuous data for abstract reasoning,'' 2013.

\bibitem{finn2017deep}
C.~Finn and S.~Levine, ``Deep visual foresight for planning robot motion,'' in
  \emph{{IEEE} Int. Conf. Robot. Autom.}, 2017, pp. 2786--2793.

\bibitem{chen2020simclear}
T.~Chen, S.~Kornblith, M.~Norouzi, and G.~Hinton, ``A simple framework for
  contrastive learning of visual representations,'' in \emph{International
  conference on machine learning}, 2020, pp. 1597--1607.

\bibitem{hotelling1933analysis}
H.~Hotelling, ``Analysis of a complex of statistical variables into principal
  components.'' \emph{J. Educational Psychology}, vol.~24, no.~6, p. 417, 1933.

\bibitem{kramer1991nonlinear}
M.~A. Kramer, ``Nonlinear principal component analysis using autoassociative
  neural networks,'' \emph{AIChE journal}, vol.~37, no.~2, pp. 233--243, 1991.

\bibitem{higgins2016beta}
I.~Higgins, L.~Matthey, A.~Pal, C.~Burgess, X.~Glorot, M.~Botvinick,
  S.~Mohamed, and A.~Lerchner, ``beta-vae: Learning basic visual concepts with
  a constrained variational framework,'' 2016.

\bibitem{goroshin2015unsupervised}
R.~Goroshin, J.~Bruna, J.~Tompson, D.~Eigen, and Y.~LeCun, ``Unsupervised
  feature learning from temporal data,'' \emph{arXiv preprint
  arXiv:1504.02518}, 2015.

\bibitem{hadsell2006dimensionality}
R.~Hadsell, S.~Chopra, and Y.~LeCun, ``Dimensionality reduction by learning an
  invariant mapping,'' in \emph{IEEE Comput. Soc. Conf. Comput. Vis. Pattern
  Recognit.}, vol.~2, 2006, pp. 1735--1742.

\bibitem{sohn2016xentloss}
K.~Sohn, ``Improved deep metric learning with multi-class n-pair loss
  objective,'' in \emph{Adv. Neural Inf. Process. Syst.}, 2016, pp. 1857--1865.

\bibitem{mcinnes2017hdbscan}
L.~McInnes, J.~Healy, and S.~Astels, ``hdbscan: Hierarchical density based
  clustering,'' \emph{J. Open Source Software}, vol.~2, no.~11, p. 205, 2017.

\bibitem{rosenberg2007v}
A.~Rosenberg and J.~Hirschberg, ``V-measure: A conditional entropy-based
  external cluster evaluation measure,'' in \emph{Joint Conf. Empirical Methods
  in Natural Language Process. and Computational Natural Language Learn.},
  2007, pp. 410--420.

\bibitem{rousseeuw1987silhouettes}
P.~J. Rousseeuw, ``Silhouettes: a graphical aid to the interpretation and
  validation of cluster analysis,'' \emph{J. Computational and Applied
  Mathematics}, vol.~20, pp. 53--65, 1987.

\bibitem{unitygameengine}
J.~K. Haas, ``A history of the unity game engine,'' 2014.

\bibitem{chamzas2020state}
C.~Chamzas, M.~Lippi, M.~C. Welle, A.~Varava, A.~Marino, L.~E. Kavraki, and
  D.~Kragic, ``State representations in robotics: Identifying relevant factors
  of variation using weak supervision,'' in \emph{Robot Learn. Wksp. at
  NeurIPS}, 2020.

\bibitem{maaten2008visualizing}
L.~V.~D. Maaten and G.~Hinton, ``Visualizing data using t-{SNE},'' \emph{J.
  Mach. Learn. Res.}, vol.~9, pp. 2579--2605, 2008.

\bibitem{scikit-learn}
F.~Pedregosa, G.~Varoquaux, A.~Gramfort, V.~Michel, B.~Thirion, O.~Grisel,
  M.~Blondel, P.~Prettenhofer, R.~Weiss, V.~Dubourg, J.~Vanderplas, A.~Passos,
  D.~Cournapeau, M.~Brucher, M.~Perrot, and E.~Duchesnay, ``Scikit-learn:
  Machine learning in {P}ython,'' \emph{J. Mach. Learn. Res.}, vol.~12, pp.
  2825--2830, 2011.

\bibitem{tipping1999probabilistic}
M.~E. Tipping and C.~M. Bishop, ``Probabilistic principal component analysis,''
  \emph{J. the Royal Statistical Society: Series B (Statistical Methodology)},
  vol.~61, no.~3, pp. 611--622, 1999.

\bibitem{he2016deep}
K.~He, X.~Zhang, S.~Ren, and J.~Sun, ``Deep residual learning for image
  recognition,'' in \emph{IEEE Conf. Comput. Vis. Pattern Recognit.}, 2016, pp.
  770--778.

\bibitem{chicco2020siamese}
D.~Chicco, ``Siamese neural networks: An overview,'' \emph{Artificial Neural
  Networks}, pp. 73--94, 2020.

\end{thebibliography}

\section{Appendix} \label{sec:appendix}

\subsection{Simulated Box Stacking Task}
\label{app:sim_box_stacking_task}

\begin{figure}[t!]

  \centering
  \includegraphics[width=0.95\linewidth]{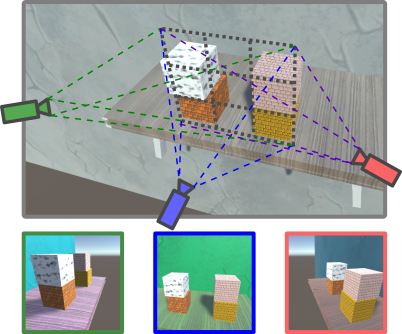}
  \caption{ Box stacking simulation task setup. Three different views are considerate.
  }\label{fig:datasets_box_stacking}
\end{figure}

\textbf{Box Stacking setup: }
The objective of this task is to plan a sequence of states leading the boxes to be stacked according to a goal image observation.
 Transitions between states, i.e., actions, can then be retrieved through the LSR~\cite{lsriros}. 
The scene, shown in \autoref{fig:datasets_box_stacking}, is composed of four boxes that can be stacked in a $3\times3$ vertical grid. The boxes are interchangeable i.e., color does not matter,  and each cell can only be occupied by one box  at a time.
The box stacking task has the following rules:
\emph{i) } a box can only be moved if there is no other box on top of it, \emph{ii) } a box can only be placed on the ground or on top of another box (but never outside the grid), \emph{iii) }no boxes can be added or completely removed from the grid (there are always $4$ boxes in play).
The arrangement of the objects in the scene represents the underlying state of the system.

Three different viewpoints are considered to capture the scene and  four datasets are built as follows:
\emph{i)}  $\mathcal{BS}_f$, where all the observations are taken from view  front  (blue in \autoref{fig:datasets_box_stacking}), \emph{ii) }
$\mathcal{BS}_r$, where the observations are only taken from view  right  (red in \autoref{fig:datasets_box_stacking}),
\emph{iii)}  $\mathcal{BS}_l$, where the observations are only taken from view  left (green in \autoref{fig:datasets_box_stacking}),  
and \emph{iv)} $\mathcal{BS}_{d}$, where we enforce that the views  for $o_i$ and $o_j$ in a training tuple are different. 
In each dataset, we randomly change lighting conditions, background, and table color in the observations. Moreover, we introduce a planar position noise of $\approx17$\% on the position of each box. 
Note that all of these changes are irrelevant factors of variation i.e., do not change the underlying system state for the box stacking task. 
 For each dataset, we use $2500$ data samples for training, with $1598$ action pairs.

%\begin{wrapfigure}{rt}{0.5\textwidth}
%\end{wrapfigure}

\begin{figure*}[htb!]
  \centering
  \includegraphics[width=\linewidth]{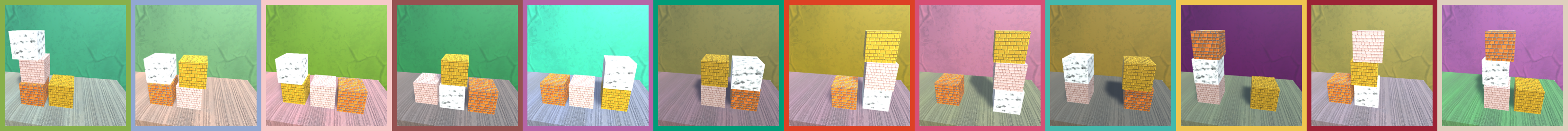}
  \caption{Example observations of the $12$ different possible states of the system. Note that the color of the boxes does not matter, i.e. boxes are interchangeable. 
  }
  \label{fig:states_12}
\end{figure*}

\noindent
\textbf{Underlying states: } Given the box stacking rules,  we can determine all the possible underlying states of the system. In particular, each state is given by a possible grid configuration specifying for every cell whether it is occupied by a box or not. Given the grid size and stacking rules, there are exactly $12$ valid different box placements are shown in \autoref{fig:states_12} and $24$ legal state transitions. 

\noindent
\textbf{Result and Discussion:}  We describe the results of this additional simulated box stacking task in the following section. In particular, we are interested to supplement the answer to the question from~\autoref{sec:exp}: {\it Do contrastive-based losses outperform reconstruction-based losses when task-irrelevant factors of variations are present in the observations?}.

\begin{figure*}[t]
  \centering
  \includegraphics[width=0.9\textwidth]{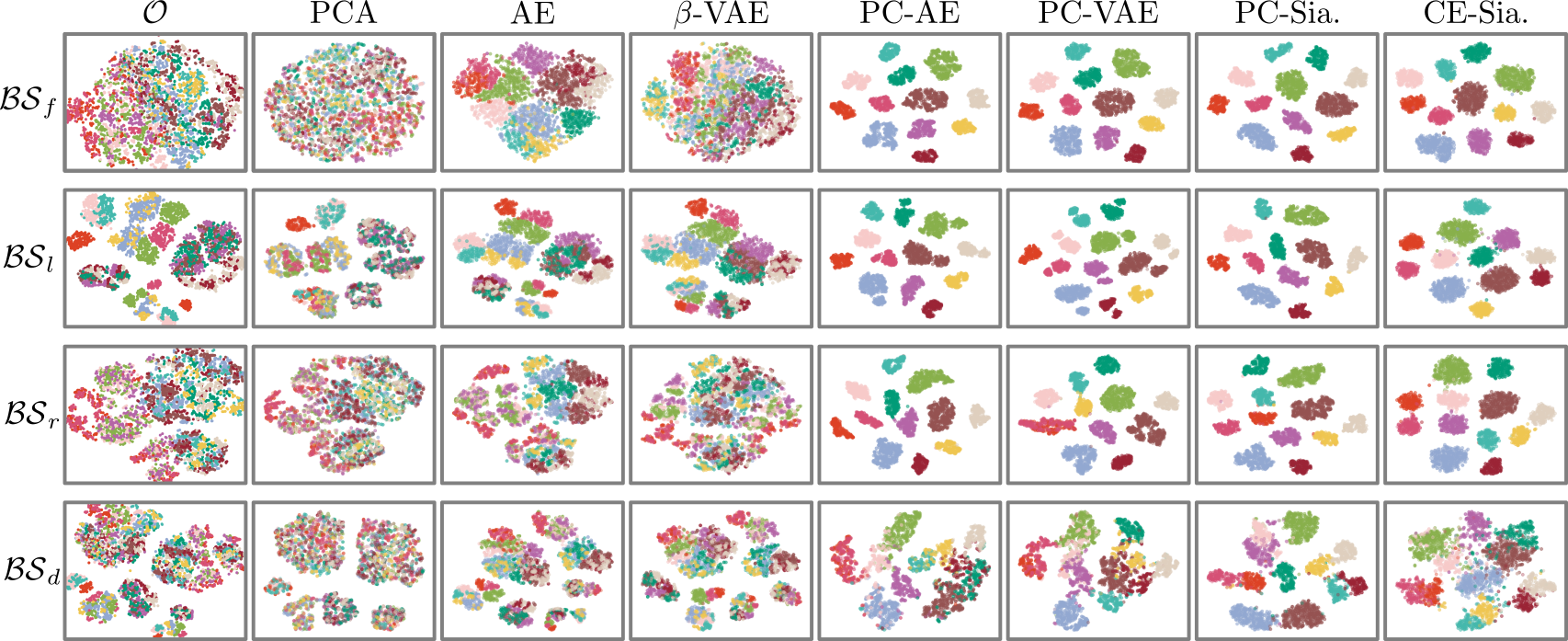}
  \caption{% Obtained $12$-dimensional representations from all models visualized in $2$D using 
  Two-dimensional t-SNE plots of the latent representations from all models for the box stacking task. The rows show  the results of the front view dataset ($\mathcal{BS}_f$), the left view ($\mathcal{BS}_l$), right view ($\mathcal{BS}_r$) and the  different viewpoints dataset ($\mathcal{BS}_d$). } 
  \label{fig:t-snes_results_box}
\end{figure*}

Similar to the other two tasks, (box manipulation and shelf arrangement task), we carry out both a qualitative and quantitative analysis on the box stacking task. Regarding the qualitative analysis, we visualize the latent encodings obtained with the different models using the respective $2$D t-SNE~\cite{maaten2008visualizing} plots in \autoref{fig:t-snes_results_box} for the box stacking task.  

 The top row shows the results from $\mathcal{BS}_f$, containing pairs taken from the frontal viewpoint. The second and third rows show the results with left $\mathcal{BS}_l$ and right $\mathcal{BS}_r$ views, respectively. Note that retrieving the system state from the observations in these datasets is more complex than in the frontal viewpoint dataset $\mathcal{BS}_f$ since the boxes occupy different sized portions of the image depending on their location.  Finally,  the last row reports the results on the dataset $\mathcal{BS}_d$, where having observations taken from different viewpoints,  making it extremely difficult to relate any visual changes to underlying state changes. 

We can observe that raw observations as well as  PCA fail completely in structuring the representations on all datasets. The purely reconstruction-based AE is able to exploit the visual similarity in the front view dataset (top row) to a certain extent structuring the encodings in a promising manner, even if not exactly separating different underlying states. However, this model clearly fails in the other datasets where few spurious clusters are formed in which the different states are mixed up.  Similar results are obtained with the $\beta$-VAE, which shows a less structured latent space for the dataset $\mathcal{BS}_f$. 
 A significant improvement of the latent space structure is recorded by introducing the PC loss term to the AE and $\beta$-VAE, 
 leading to $12$ visually distinct clusters for $\mathcal{BS}_f$, $\mathcal{BS}_l$ and $\mathcal{BS}_r$ (corresponding to the $12$ underlying states of the system).  
The same applies to both purely contrastive loss-based Siamese networks (PC-Siamese and CE-Siamese). The different viewpoint dataset however is clearly more challenging for all models. %

 The qualitative observations based on t-SNE plots are confirmed by the numerical analysis reported in Table~\ref{tab:results_all_box_stacking}. 
 %In detail,  we can 
 We observe that all models employing a contrastive loss (PC-AE, PC-VAE, PC-Siamese, and CE-Siamese) perform nearly perfectly for the frontal view dataset $\mathcal{BS}_f$ (top left) in regard to all the performance metrics: the recorded homogeneity ($h_c$) and completeness ($c_c$)  show a perfect clusterability and therefore enable the LSR to successfully plan in $100\%$ of the cases (the exception being the CE-Siamese with $98.8\%$ which indicates that only a start or goal state was misclassified).  In contrast,  directly using the raw observation, the PCA or  reconstruction-based models achieve very low clustering and planning performance, reaching values $<5\%$ for the existence of at least one correct path.

 As far as the left viewpoint dataset is concerned (top right), all the models not employing any form of contrastive loss (PCA, AE, $\beta$-VAE) achieve low clustering (i.e., low $h_c,c_c,s_c$) and planning performance (i.e., low $c_e$, \textit{\% any} and \textit{\% all} with $\mathcal{V}\neq 12$ and $\mathcal{E}\neq 24$).  The results show that the best performing model is the PC-AE that achieves $100\%$ for \textit{\% all} and \textit{\% any} in the planning scores. It manages to build $16$ clusters (only  $4$  clusters more than the ideal number) that have perfect homogeneity,  completeness and a good silhouette score ($0.85$). 
 
The other purely contrastive-based models (PC-Siamese and CE-Siamese) achieve both lower performance of $65-70$\% and fall behind the PC-VAE with $76$\% planning performance for \textit{\% any} path. 
This is explained by imprecise clustering (shown by sub-optimal homogeneity scores $\approx 0.9$) that compromises the planning performance.

\begin{table*}[ht!]
\resulttable{$\mathcal{BS}_f$}{
- & $118$ & $0.61$ & $0.62$ & $0.29$ & $119$ & $0.78$ & $2.53$ & $2.8$
\\ \hline
PCA &  $107$ & $0.67$ & $0.72$ & $0.33$ & $123$ & $0.79$ & $4.04$ & $4.20$
\\ \hline 
AE &  $28$ & $0.85$ & $0.87$ & $0.38$ & $19$ & \boldmath$1.00$ & $2.60$ & $2.60$
\\ \hline 
$ \beta$-VAE  & $52$ & $0.8$ & $0.81$ & $0.36$ & $38$ & $0.82$ & $1.60$ & $1.60$
\\ \hline 
PC-AE &  $23$ & \boldmath$1.00$ & \boldmath$1.00$ & \boldmath$0.86$ & \boldmath$59$ & \boldmath$1.00$ & \boldmath$100.0$ & \boldmath$100.0$
\\ \hline 
PC-VAE &  $19$ & \boldmath$1.00$ & \boldmath$1.00$ & $0.84$ & $48$ & \boldmath$1.00$ & \boldmath$100.0$ & \boldmath$100.0$
\\ \hline 
PC-Sia. &  $18$ & \boldmath$1.00$ & \boldmath$1.00$ & $0.84$ & $44$ & \boldmath$1.00$ & \boldmath$100.0$ & \boldmath$100.0$
\\ \hline 
CE-Sia. &  \boldmath$13$ & \boldmath$1.00$ & \boldmath$1.00$ & $0.35$ & $28$ & $0.96$ & $98.80$ & $98.80$
\\ \hline 
%%%%%%%%%%%%%%%%%%%%%  
 \hline
&  \multicolumn{8}{c|}{Dataset  $\mathcal{BS}_d$}
\\ \hline 
- &  $92$ & $0.15$ & $0.24$ & $0.28$ & $157$ & $0.61$ & $0.95$ & $1.0$
\\ \hline
PCA &  $40$ & $0.03$ & $0.08$ & $0.16$ & $91$ & $0.63$ & $2.65$ & $2.70$
\\ \hline 
AE &  \boldmath $21$ & $0.04$ & $0.12$ & $0.13$ &  \boldmath$43$ & $0.58$ & $1.07$ & $1.20$
\\ \hline 
$ \beta$-VAE  &  $29$ & $0.05$ & $0.10$ & $0.05$ & $49$ & $0.61$ & $2.17$ & $2.40$
\\ \hline 
PC-AE &  $80$ & $0.92$ & $0.92$ & $0.26$ & $136$ & $0.89$ & \boldmath $59.61$ & $63.20$
\\ \hline 
PC-VAE &  $43$ & $0.70$ & $0.93$ & $0.05$ & $92$ & $0.74$ & $30.70$ & $33.70$
\\ \hline 
PC-Sia. &  $28$ & \boldmath $0.94$ & \boldmath $0.99$ & \boldmath $0.48$ & $60$ & $0.83$ & $57.23$ & \boldmath$ 66.30$
\\ \hline 
CE-Sia. &  $37$ & $0.89$ & $0.97$ & $0.19$ & $66$ & \boldmath $0.92$ & $54.38$ & $59.40$
\\ \hline  
 }
\resulttable{$\mathcal{BS}_l$}{
- &  $131$ & $0.34$ & $0.53$ & $0.35$ & $192$ & $0.68$ & $5.4$ & $5.6$
\\ \hline
PCA &  $85$ & $0.32$ & $0.51$ & $0.27$ & $154$ & $0.75$ & $13.74$ & $15.90$
\\ \hline 
AE &  $33$ & $0.28$ & $0.52$ & $0.12$ & \boldmath$33$ & $0.67$ & $8.90$ & $8.90$
\\ \hline 
$ \beta$-VAE  &  $29$ & $0.44$ & $0.60$ & $0.05$ & $35$ & $0.66$ & $3.50$ & $3.50$
\\ \hline 
PC-AE & \boldmath $16$ & \boldmath $1.00$ & \boldmath $1.00$ & \boldmath $0.85$ & $38$ & \boldmath $1.00$ & \boldmath $100.0$ & \boldmath $100.0$
\\ \hline 
PC-VAE &  $18$ & $0.90$ & \boldmath $1.00$ & $0.75$ & $40$ & \boldmath $1.00$ & $76.30$ & $76.30$
\\ \hline 
PC-Sia. &  $17$ & $0.91$ & \boldmath $1.00$ & $0.72$ & $39$ & $0.87$ & $56.67$ & $64.20$
\\ \hline 
CE-Sia. & \boldmath $16$ & $0.92$ & \boldmath $1.00$ & $0.33$ & $39$ & $0.87$ & $61.40$ & $69.20$
\\ \hline 
%%%%%%%%%%%%%%%%%%%%% 
 \hline
&  \multicolumn{8}{c|}{Dataset  $\mathcal{BS}_r$} 
\\ \hline
- &  $123$ & $0.67$ & $0.71$ & $0.3$ & $124$ & $0.69$ & $0.6$ & $0.6$
\\ \hline
PCA &  $119$ & $0.54$ & $0.56$ & $0.33$ & $151$ & $0.62$ & $2.30$ & $2.80$
\\ \hline 
AE &  $35$ & $0.71$ & $0.74$ & $0.33$ & \boldmath$24$ & $0.88$ & $1.10$ & $1.10$
\\ \hline 
$ \beta$-VAE  &  $41$ & $0.76$ & $0.79$ & $0.37$ & $32$ & $0.81$ & $1.20$ & $1.20$
\\ \hline 
PC-AE &  \boldmath $21$ &\boldmath$1.00$ & \boldmath $1.00$ & \boldmath $0.74$ & $54$ & \boldmath $0.96$ & \boldmath $90.70$ & \boldmath $95.40$
\\ \hline 
PC-VAE &  \boldmath $21$ & $0.93$ & \boldmath $1.00$ & $0.59$ & $53$ & $0.91$ & $68.50$ & $73.90$
\\ \hline 
PC-Sia. &  $22$ & $0.96$ &\boldmath $1.00$  & $0.63$ & $54$ & $0.91$ & $73.56$ & $80.80$
\\ \hline 
CE-Sia. &  $27$ & $0.97$ &\boldmath $1.00$ & $0.31$ & $63$ & $0.92$ & $81.70$ & $83.70$
\\ \hline
}
\caption{
 Evaluation results according to~\autoref{sec:scores} for the mapping models (rows $2-8$ of each table) and the raw observation (first row of each table) on datasets $\mathcal{BS}_f$ and $\mathcal{BS}_{d}$ on the left, and   $\mathcal{BS}_l$ and $\mathcal{BS}_{r}$ on the right for the simulated box stacking task. Best results in bold.}
\label{tab:results_all_box_stacking}
\end{table*}

\begin{figure*}[ht!]
  \centering
  \includegraphics[width=\linewidth]{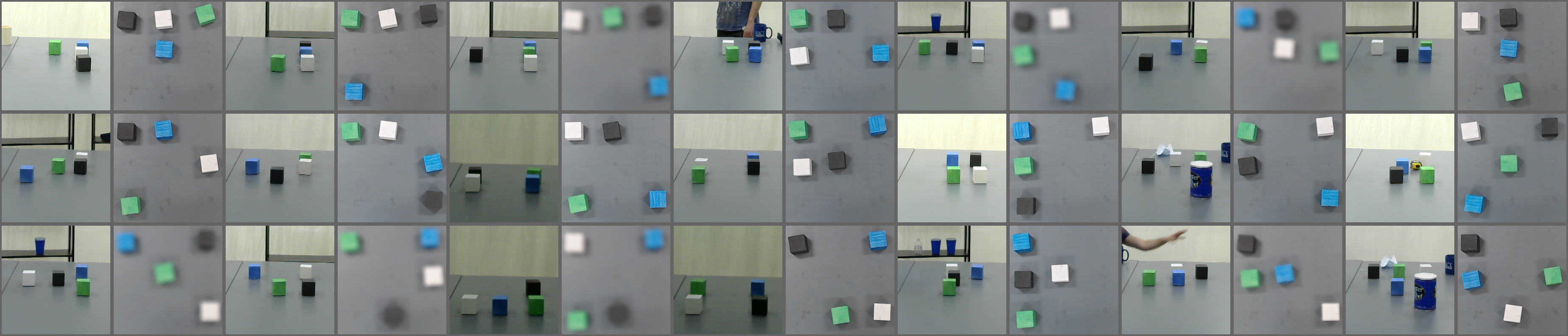}
  \caption{$70/126$ example observation from the side and top view for the box manipulation task.
  }
  \label{fig:states_real_push}
\end{figure*}
 
Similar considerations also hold for results of the right viewpoint dataset in 
\autoref{tab:results_all_box_stacking} (bottom right). More specifically,  poor clustering and planning metrics are achieved by the models that do not rely on the contrastive loss, while best general performance is obtained with PC-AE, which reaches $95.4\%$ for \textit{\% any} as well as best clustering scores ($h_c=c_c=1$ and $s_c=0.74$). Acceptable planning performance is also achieved by the other models using the contrastive loss ($\approx 73-83$\% for \textit{\%~any}).   

However, note that the good performance of the PC-AE does not translate to the dataset where different views are enforced ($\mathcal{BS}_d$) (bottom left). This suggests that it is able to combine reconstruction and contrastive losses beneficially as long as all the observations are obtained with a fixed viewpoint,  but fails when the differences in observation become too large.  In contrast,  the purely contrastive models can effectively handle such differences. In this regard, it can be noticed that PC-AE, PC-Siamese, and CE-Siamese achieve the best (comparable) performance in terms of clustering and planning.
Interestingly, the PC-AE is able to combine the reconstruction loss with the additional contrastive loss more effectively than the PC-VAE model. Possibly this is due to a better PC loss coefficient $\alpha$ choice or to the absence of the KL-term. Finally, on the performance of the other models on dataset $\mathcal{BS}_d$, we observe poor clustering and planning performance when using raw observations, PCA, AE, and $\beta$-VAE, confirming the relevance of  task priors
%weak supervision 
to handle task-irrelevant factors of variation.  

\textit{In summary,  we observe that models involving the contrastive term (PC-AE, PC-VAE, PC-Siamese, CE-Siamese) 
outperform the ones that do not (PCA, AE, $\beta$-VAE) by a significant margin.}

\begin{figure*}[ht!]
  \centering
  \includegraphics[width=\linewidth]{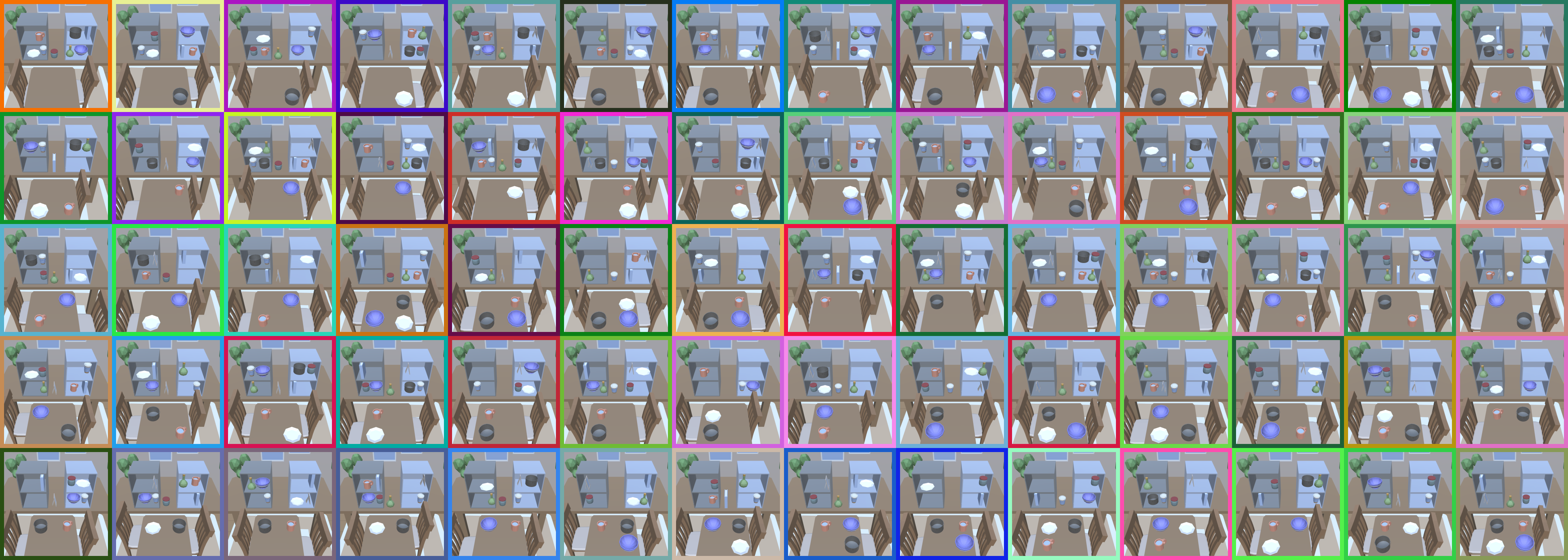}
  \caption{Example observations of the $70$ different possible states of the self arraignment system. Note that the individual object does not matter, i.e. objects are interchangeable. 
  }
  \label{fig:states_70}
\end{figure*}

%------------------------------------------------------------------------------------------------------------------
%----------------- Dataset and underlying state explanations
%------------------------------------------------------------------------------------------------------------------
\subsection{Box Manipulation Setup and Underlying States}
\label{app:pushing_rules}

\noindent
\textbf{Box Manipulation setup: } The box manipulation task involves $4$ boxes which are arranged in a predetermined $3\times3$ grid. % where each cell can only be occupied by one box at a time. 
The rules for the task are the following: \emph{i)} a box can not be moved into a cell that already contains a box, \emph{ii)} a box can only be moved into the four cardinal directions (as it happens in a box pushing task), \emph{iii)} a box can only be moved one cell at the time.
A similar
%no-action 
pair is done by swapping the position of boxes and an action pair by performing a random move.  Before executing an action we check its validity (preconditions)  as described in~\ref{sec:val-setting}.
% {\red by placing the gripper in each affecting cell and checking if an object is in the cell. This is done by closing the gripper and checking if the gripper fully closed (no object) or did not fully close (an object exists).
% For example, for action pairs, one cell must be occupied and the next cell empty, while for  similar %no action-
% pairs both cells must be occupied. This allows the robot to collect data in a self-supervised manner.
% Since discovering valid actions and safely executing them is not the scope of this paper this simple procedure is not shown in the supplementary videos.}

During the collection, a number of task-irrelevant objects (hat, coffee-mug, mask, etc.) are placed in the view field of the side camera and people are moving in the background along with several other objects. The top view often performs auto focus when the robotic arm is moving the boxes which lead to blurry observations. Example observations are shown in \autoref{fig:states_real_push}.

 In total $2135$ training data pairs were collected in a self-supervised manner in $\approx$ 30 hours.
 As the action between states are reversible, we also included the reversed version of each action pair, i.e., given the tuple $(o_1,o_2, s= 0)$ we add $(o_2,o_1, s= 0)$.

\noindent
\textbf{Underlying states: } Given the box manipulation rules,  we can determine all the possible underlying states of the system. In particular, each state is given by a possible grid configuration specifying for every cell whether it is occupied by a box or not. Given the grid size and manipulation rules, there are exactly $126$ different box placements, i.e. grid configurations.

\subsection{Shelf Arrangement Setup and Underlying States}
\label{app:stacking_rules_shelf}

\noindent
\textbf{Shelf arrangement setup:}
The shelf arrangement setup is composed of a table and two shelves. The table has four potential slots for  task relevant objects. For the shelves, each shelf slot can be occupied at most by one task relevant object. The task object can however be placed either to the left or the right inside the shelf itself. The task relevant objects are always present in the scene. An action moves a relevant object from the table to the shelf or vice-versa while a swapping motion (a similar pair) exchanges the position of two objects. A small amount of positional noise for the objects is also introduced each time the scene is generated.
Five distractor objects, that are not relevant for the task,  can be present in the shelf slots. Their position inside the shelf slots is not fixed. 

\noindent
\textbf{Underlying states:}
The rules for the shelf arrangement task result in a system that has exactly $70$ underlying distinct states  and 320 legal transitions, which are obtained by considering all possible combinations of object configurations in the $8$ available slots.
\autoref{fig:states_70} shows examples of all $70$ distinct underlying states for the dataset with no distractors present.

\begin{table*}[ht!]
%%%%%%%%%%%%%%%%%%%%% rand 1 act 1 augmentation 
\resulttable{
${{\mathcal{BM}}}_t$ }{ 
- & $1016$ & $0.92$ & $0.92$ & $0.79$ & $583$ & $0.78$ & $0.0$ & $0.0$
\\ \hline 
PCA &  $496.0$ & $0.75$ & $0.78$ & $0.52$ & $452.0$ & $0.52$ & $0.0$ & $0.0$
\\ \hline 
AE &  $233.0$ & $0.49$ & $0.57$ & $0.29$ & $234.0$ & $0.27$ & $0.0$ & $0.0$
\\ \hline 
$ \beta$-VAE  &  $539.0$ & $0.85$ & $0.85$ & $0.51$ & \boldmath$422.0$ & $0.62$ & $0.0$ & $0.0$
\\ \hline 
PC-AE &  $246.0$ & $0.54$ & $0.6$ & $0.3$ & $258.0$ & $0.28$ & $0.0$ & $0.0$
\\ \hline 
PC-VAE &  $570.0$ & \boldmath$1.0$ & \boldmath$1.0$ & $0.58$ & $488.0$ & \boldmath$1.0$ & $29.9$ & $29.9$
\\ \hline 
PC-Sia. &  $389.0$ & $0.99$ & \boldmath$1.0$ & $0.52$ & $458.0$ & $0.97$ & $47.37$ & $57.3$
\\ \hline 
CE-Sia. &  \boldmath$150.0$ & \boldmath$1.0$ & \boldmath$1.0$ & \boldmath$0.67$ & $325.0$ & \boldmath$1.0$ & \boldmath$98.3$ & \boldmath$98.3$
\\ \hline
%%%%%%%%%%%%%%%%%%%%% rand 2 act 1 augmentation 
 \hline
&  \multicolumn{8}{c|}{Dataset  $\overline{\mathcal{BM}}_t$}
\\ \hline
PCA &  - & - & - & - & - & - & - & - %$502.0$ & $0.74$ & $0.77$ & $0.52$ & $451.0$ & $0.5$ & $0.0$ & $0.0$
\\ \hline 
AE & - & - & - & - & - & - & - & - % $219.0$ & $0.46$ & $0.55$ & $0.27$ & $238.0$ & $0.19$ & $0.0$ & $0.0$
\\ \hline 
$ \beta$-VAE  & - & - & - & - & - & - & - & - % $538.0$ & $0.83$ & $0.84$ & $0.52$ & \boldmath$432.0$ & $0.53$ & $0.0$ & $0.0$
\\ \hline 
PC-AE &  $218.0$ & $0.99$ & \boldmath$1.0$ & $0.71$ & $375.0$ & $0.98$ & $72.12$ & $82.5$
\\ \hline 
PC-VAE &  $395.0$ & \boldmath$1.0$ & \boldmath$1.0$ & $0.56$ & $461.0$ & \boldmath$1.0$ & $89.7$ & $89.7$
\\ \hline 
PC-Sia. &  \boldmath$133.0$ & \boldmath$1.0$ & \boldmath$1.0$ & \boldmath$0.9$ & $314.0$ & \boldmath$1.0$ & \boldmath$97.7$ & \boldmath$98.2$
\\ \hline 
CE-Sia. &  - & - & - & - & - & - & - & -
\\ \hline 
}
\resulttable{ $\mathcal{BM}_s$}{
- &  $1002$ & $0.96$ & $0.96$ & $0.78$ & $558$ & $0.89$ & $0.0$ & $0.0$
\\ \hline
PCA &  $539.0$ & $0.74$ & $0.75$ & \boldmath$0.59$ & $479.0$ & $0.35$ & $0.0$ & $0.0$
\\ \hline 
AE &  $532.0$ & $0.77$ & $0.78$ & $0.57$ & $459.0$ & $0.41$ & $0.0$ & $0.0$
\\ \hline 
$ \beta$-VAE  &  $531.0$ & $0.79$ & $0.8$ & \boldmath$0.59$ & $456.0$ & $0.45$ & $0.0$ & $0.0$
\\ \hline 
PC-AE &  $550.0$ & $0.79$ & $0.8$ & \boldmath$0.59$ & $456.0$ & $0.47$ & $0.0$ & $0.0$
\\ \hline 
PC-VAE &  $548.0$ & \boldmath$0.99$ & \boldmath$0.99$ & $0.57$ & $460.0$ & \boldmath$0.98$ & $10.0$ & $10.0$
\\ \hline 
PC-Sia. &  $345.0$ & $0.95$ & $0.98$ & $0.48$ & $391.0$ & $0.91$ & $31.02$ & $34.2$
\\ \hline 
CE-Sia. &  \boldmath$298.0$ &\boldmath$0.99$ & \boldmath$0.99$ & $0.41$ & \boldmath$407.0$ & $0.94$ & \boldmath$71.57$ & \boldmath$72.2$
\\ \hline 
%%%%%%%%%%%%%%%%%%%%% rand 2 act 0 augmentation 
 \hline
&  \multicolumn{8}{c|}{Dataset  $\overline{\mathcal{BM}}_s$}
\\ \hline
PCA & - & - & - & - & - & - & - & - % $534.0$ & $0.74$ & $0.75$ & \boldmath$0.59$ & $474.0$ & $0.37$ & $0.0$ & $0.0$
\\ \hline 
AE & - & - & - & - & - & - & - & - % $521.0$ & $0.79$ & $0.79$ & $0.57$ & $443.0$ & $0.42$ & $0.0$ & $0.0$
\\ \hline 
$ \beta$-VAE  &  - & - & - & - & - & - & - & - %$525.0$ & $0.8$ & $0.81$ & $0.56$ & \boldmath$438.0$ & $0.42$ & $0.0$ & $0.0$
\\ \hline 
PC-AE &  $420.0$ & $0.98$ & $0.99$ & $0.53$ & $456.0$ & $0.95$ & $40.89$ & $44.9$
\\ \hline 
PC-VAE &  $519.0$ & \boldmath$1.0$ & \boldmath$1.0$ & $0.57$ & $448.0$ & \boldmath$0.99$ & $31.5$ & $31.5$
\\ \hline 
PC-Sia. &  \boldmath$304.0$ & $0.97$ & $0.99$ & $0.5$ & $394.0$ & $0.92$ & \boldmath$46.6$ & \boldmath$55.6$
\\ \hline 
CE-Sia. &  - & - & - & - & - & - & - & -
\\ \hline 
}
\resulttable{ $\mathcal{BM}_{st}$}{
- & $710$ & $0.83$ & $0.83$ & $0.5$ & $496$ & $0.58$ & $0.0$ & $0.0$
\\ \hline
PCA &  $400.0$ & $0.59$ & $0.62$ & $0.46$ & \boldmath$453.0$ & $0.25$ & $0.0$ & $0.0$
\\ \hline 
AE &  $554.0$ & $0.87$ & $0.88$ & $0.56$ & $454.0$ & $0.69$ & $0.0$ & $0.0$
\\ \hline 
$ \beta$-VAE  &  $407.0$ & $0.72$ & $0.74$ & $0.44$ & $361.0$ & $0.38$ & $0.0$ & $0.0$
\\ \hline 
PC-AE &  $318.0$ & $0.62$ & $0.91$ & \boldmath$0.61$ & $325.0$ & $0.47$ & $0.0$ & $0.0$
\\ \hline 
PC-VAE &  $381.0$ & $0.84$ & $0.85$ & $0.42$ & $295.0$ & $0.65$ & $0.1$ & $0.1$
\\ \hline 
PC-Sia. &  $289.0$ & $0.96$ & $0.96$ & $0.4$ & $312.0$ & $0.92$ & $26.34$ & $27.5$
\\ \hline 
CE-Sia. &  \boldmath$232.0$ & \boldmath$0.99$ & \boldmath$0.99$ & $0.41$ & $354.0$ & \boldmath$0.99$ & \boldmath$78.39$ & \boldmath$78.7$
\\ \hline 
%%%%%%%%%%%%%%%%%%%%% rand 2 act 0 augmentation 
 \hline
&  \multicolumn{8}{c|}{Dataset  $\overline{\mathcal{BM}}_{st}$} 
\\ \hline
PCA & - & - & - & - & - & - & - & - % $404.0$ & $0.61$ & $0.63$ & $0.47$ & $436.0$ & $0.25$ & $0.0$ & $0.0$
\\ \hline 
AE & - & - & - & - & - & - & - & - % $562.0$ & $0.86$ & $0.87$ & \boldmath$0.55$ & $454.0$ & $0.68$ & $0.1$ & $0.1$
\\ \hline 
$ \beta$-VAE  & - & - & - & - & - & - & - & - % $409.0$ & $0.73$ & $0.74$ & $0.43$ & $351.0$ & $0.39$ & $0.0$ & $0.0$
\\ \hline 
PC-AE &  $383.0$ & $0.96$ & $0.98$ & $0.53$ & \boldmath$433.0$ & $0.88$ & $40.77$ & $48.9$
\\ \hline 
PC-VAE &  $335.0$ & $0.91$ & $0.92$ & $0.42$ & $274.0$ & $0.84$ & $0.7$ & $0.7$
\\ \hline 
PC-Sia. &  \boldmath$235.0$ & \boldmath$0.99$ & \boldmath$0.99$ & $0.41$ & $337.0$ & \boldmath$0.99$ & \boldmath$77.7$ & \boldmath$78.8$
\\ \hline 
CE-Sia. &  - & - & - & - & - & - & - & -
\\ \hline 
}
\caption{ 
Evaluation results of the models for the box manipulation task. Datasets  $\mathcal{BM}_{t}$ (top left part), $\mathcal{BM}_{s}$ (top right part), and $\mathcal{BM}_{t}$ (bottom left part)  with respective augmented versions are reported. Best results in bold.   }
\label{tab:results_all_box_manip}
\end{table*}

\subsection{Architectures and Hyperparameters}

\label{app:architecture}
In this section, we describe the architectural details, as well as the hyperparameters for all considered models. The input dimension for all models is a  $256\times256\times3$ image. To make the comparison between the models as fair as possible, each uses the same training data and latent space dimension.

\noindent
\textbf{PCA:}
We used the popular scikit-learn~\cite{scikit-learn} implementation of the principal component analysis based on~\cite{tipping1999probabilistic}. The number of components was set to $12$ and the model fit to the training dataset. The dimension reduction was then applied to the holdout dataset for evaluation.

\noindent
\textbf{AE:}
The implementation of encoder and decoder for the Auto-encoder~\cite{kramer1991nonlinear} relies on the ResNet architecture in~\cite{he2016deep}, with a depth of two per block for the box stacking and manipulation tasks and a larger ResNet architecture having six layers with depth two for the shelf stacking task. We train each model for $500$ epochs and a batch size of $64$.

\noindent
\textbf{$\beta$-VAE: } 
   The implementation of encoder and decoder of the $\beta$-VAE \cite{higgins2016beta} is realized by adding the probabilistic components to the AE architecture. 
   We train the models for $500$ epochs with a scheduling for beta from $0$ to $1.5$ and a batch size of $64$.   

\noindent
\textbf{PC-AE:}
The PC-AE uses the same architecture as the AE model.

\noindent
\textbf{PC-VAE: }  
The PC-VAE uses the same architecture as the $\beta$-VAE model. 

\noindent
 \textbf{PC-Siamese:} 

The Siamese network \cite{chicco2020siamese}  architecture is comprised of two identical encoder networks. Each encoder has the following latent encoding  architecture:
\begin{align*}
x_1 &=  MaxPool(x, 2\times2) \\
x_2 &=  Conv(x_1, 4\times4, relu) \\
x_3 &=  Conv(x_2, 4\times4, relu) \\
x_4 &=  MaxPool(x_3, 7\times7) \\
z  & =  FC(x_4, 12, relu)
\end{align*}

\noindent
\textbf{CE-Siamese:}
 This model has the same architecture as the Siamese but uses the normalized temperature-scaled cross entropy loss.
 
\noindent
\textbf{Hyperparameters: }  The latent space dimension was set to $12$ for all $56$ models.
 Concerning the hyperparameters in the loss functions,  we employed the same scheduling as in \cite{lsriros} for  $\alpha$ and $\gamma$ 
 for the losses employed in PC-AE and PC-VAE

 and for $\beta$ in \eqref{eq:vaeloss}, reaching $\alpha = 100$, $\gamma = 2500$ and $\beta = 2$. 
 In order to set the minimum distance $d_m$ in \eqref{eq:acloss} for PC-AE and PC-VAE, we leveraged the approach  in \cite{lsriros} based on 
 measuring the average distance of the action pairs in the models AE and $\beta$-VAE, while we set it to  $0.5$ for PC-Siamese. The AE, $\beta$-VAE, PC-AE and PC-VAE where trained for $500$ epochs while the PC-Sia. and CE-Sia 
 were trained for $100$ epochs.  
 Concerning the HDBSCAN parameter, denoting the minimum number of samples in each cluster, we  set it to $5$ for the datasets obtained in simulation and to $2$ for the real-world datasets. This is motivated by the fact that a smaller amount of data is available for the real-world setting. 

%------------------------------------------------------------------------------------------------------------------
%----------------- More experiments results
%------------------------------------------------------------------------------------------------------------------

\begin{figure*}[ht]
  \centering
  \includegraphics[width=0.9\textwidth]{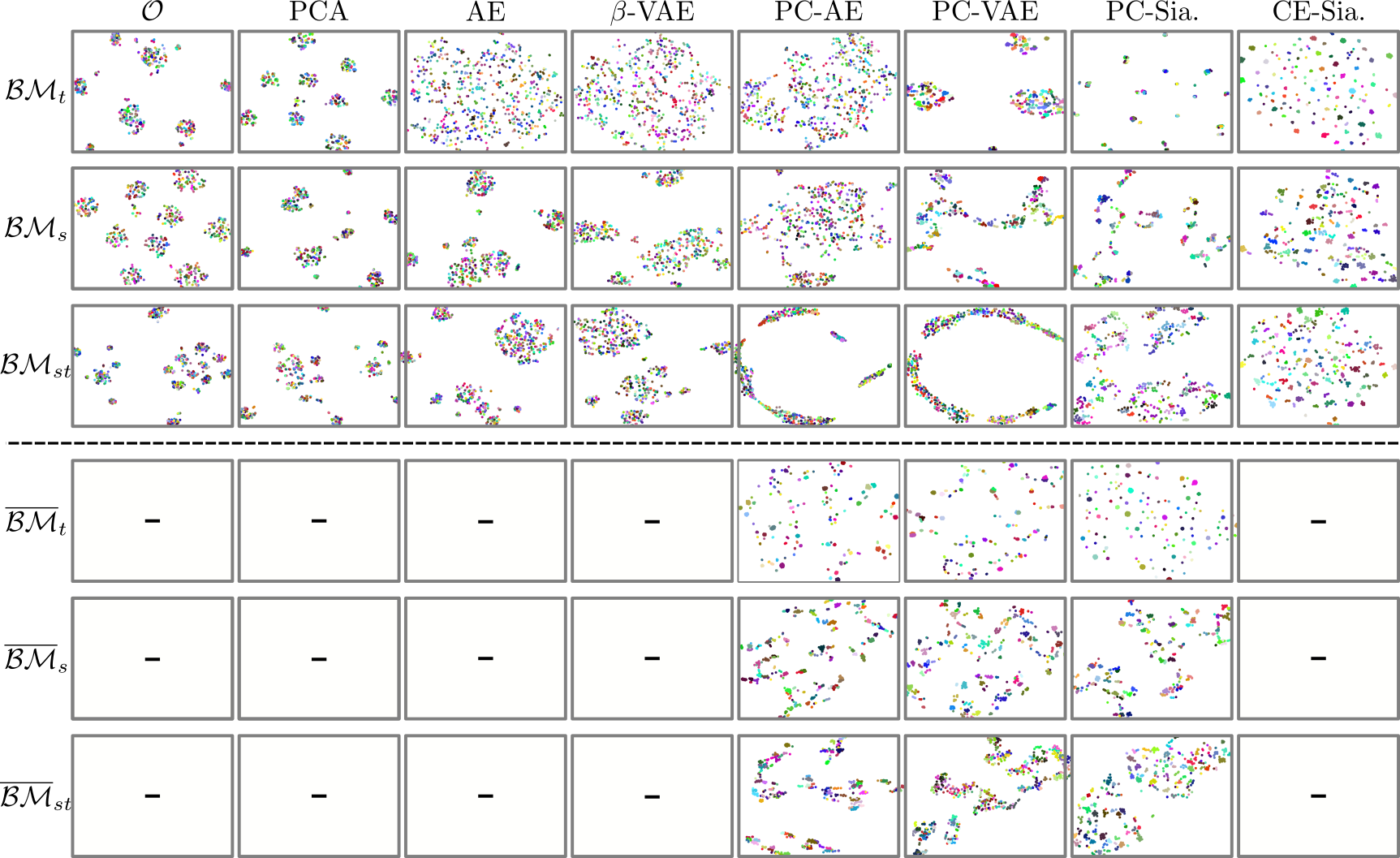}
  \caption{
Two-dimensional t-SNE plots of the latent representations from all models for the box manipulation task. The rows shows the results of the top view ($\mathcal{BM}_t$), the side view  ($\mathcal{BM}_s$),  and the mixed viewpoints dataset ($\mathcal{BM}_{st}$), as well as their augmented versions in descending order.} 
  \label{fig:t-snes_box_maniup_all}
\end{figure*}

\subsection{Additional Experimental Results for Box Manipulation}\label{sec:app:exp_box_push}
Here we mention some additional results for the box manipulation task.

\noindent
\textbf{Experimental Results:}
\autoref{tab:results_all_box_manip} shows all the results for the box manipulation task for all $7$ mapping models, as well as for the raw observations, on the top view 
($\mathcal{BM}_t$ - top left in the table), side ($\mathcal{BM}_s$ - top right in the table) and mixed view 
($\mathcal{BM}_{st}$ - bottom left in the table) datasets as well as their augmented versions  ($\overline{\mathcal{BM}}_t$, 
$\overline{\mathcal{BM}}_s$, $\overline{\mathcal{BM}}_{st}$).  We will focus the discussion on the results that were omitted in the discussion in \autoref{sec:exp}, i.e., we will focus on the side view dataset. 

Concerning the side view dataset, we observe that, as for $\mathcal{BM}_t$ and $\mathcal{BM}_{st}$, the CE-Siamese achieves overall best planning and clustering performance, reaching $72.2\%$ for \textit{\% any} and almost perfect homogeneity and completeness ($h_c=c_c=0.99$) with $298$ clusters. 
General lower performance is obtained by the other pure contrastive-based model 
PC-Siamese, which reaches $34.2\%$  for \textit{\% any} and creates a higher number of clusters ($|\mathcal{V}|=345$) which are less homogeneous and complete ($h_c = 0.5$, $c_c=0.98$). A further decrease on the planning performance is observed with PC-VAE  ($10.0\%$ for \textit{\% any}) which leads to creating a large number of clusters ($|\mathcal{V}|=548$) with good homogeneity and completeness ($h_c=c_c=0.99$). All the other models (PCA, AE, $\beta$-VAE, and PC-AE) are getting $0\%$ planning performance.

 Regarding the augmentation, we do not evaluate  the case of the raw observations or the models not using any contrastive component (PCA, AE, $\beta$-VAE) since such models do not exploit the random action pairs generated by the augmentation. Similarly, we do not  
evaluate CE-Siamese as it does not use the action pairs in the dataset.  Concerning the other models,  interestingly the augmentation propels the PC-AE from having $0\%$ for \textit{\% any} on the dataset $\mathcal{BM}_s$ to having $44.9\%$ for the augmented version $\overline{\mathcal{BM}}_s$. 
An improvement of $\approx25\%$ is also recorded for the PC-VAE and  PC-Siamese, with the latter outperforming the former ($31.5\%$ for PC-VAE, $55.6\%$ for PC-Siamese for the score \textit{\% any}). 

 The quantitative analysis is also reflected in the t-SNE plots in \autoref{fig:t-snes_box_maniup_all}. We can observe that only CE-Siamese obtains a good separation for all datasets. Moreover, the random sampling data augmentation  significantly improves the structure of the latent spaces obtained by PC-AE, PC-VAE and PC-Siamese for the top and side views,  while lower improvement is recorded for the mixed view dataset.

\subsection{Additional Experimental Results for Shelf Arrangement}\label{sec:app:exp_shelf}
Table~\ref{tab:results_all_shelf_arranging} reports the evaluation results with all the datasets in the shelf arrangement task. 
The left column reports the datasets that have no distractor present, while the right column shows the performance for the dataset where all five distractors are present (with probability $0.8$). The top row block are all models for the non-augumented datasets ($\mathcal{SA}_{0d}$ and $\mathcal{SA}_{5d}$).
The second row block indicates the datasets augmented with randomly sampled dissimilar pairs like described in \autoref{sec:problem} ($\overline{\mathcal{SA}}_{0d}$ and $\overline{\mathcal{SA}}_{5d}$).  
%
%{\red The third row block contains datasets  with  reshuffled no-action pairs ($\widetilde{\mathcal{SA}}_{0d}$ and $\widetilde{\mathcal{SA}}_{5d}$) as described in \autoref{sec:app:aug}, while the last row block shows the combination of both data augmentation methods ($\widetilde{\overline{\mathcal{SA}}}_{0d}$ and $\widetilde{\overline{\mathcal{SA}}}_{5d}$).}
%
Note that the CE-Siamese is not shown for the augmentations that only alter the dissimilar pairs as they are not used in this particular model.

Concerning the non-augmented datasets ${\mathcal{SA}}_{0d}$ and ${\mathcal{SA}}_{5d}$, also the performance of models PCA, AE and $\beta$-VAE are reported in addition to the ones in the \autoref{sec:exp} confirming that none of the non-augmented datasets (except for CE-Siamese) achieve any meaningful performance. 
 Regarding the augmented datasets, we can observe they lead to much 
better structured latent spaces for all models in the case of no distractors. Especially the 
PC-Siamese model achieves perfect clustering score for the 
homogeneity and compactness, as well as perfect planning performance of $100\%$ for both \textit{$\%$ any} and \textit{$\%$ all}. However, the augmentation  only helps to a marginal extent for the case of five distractor objects present, where only the PC-Siamese model improves to $13.1\%$ for \textit{$\%$ any}.

\begin{figure*}
%\vspace{-9cm}
  \centering
  \includegraphics[width=0.9\textwidth]{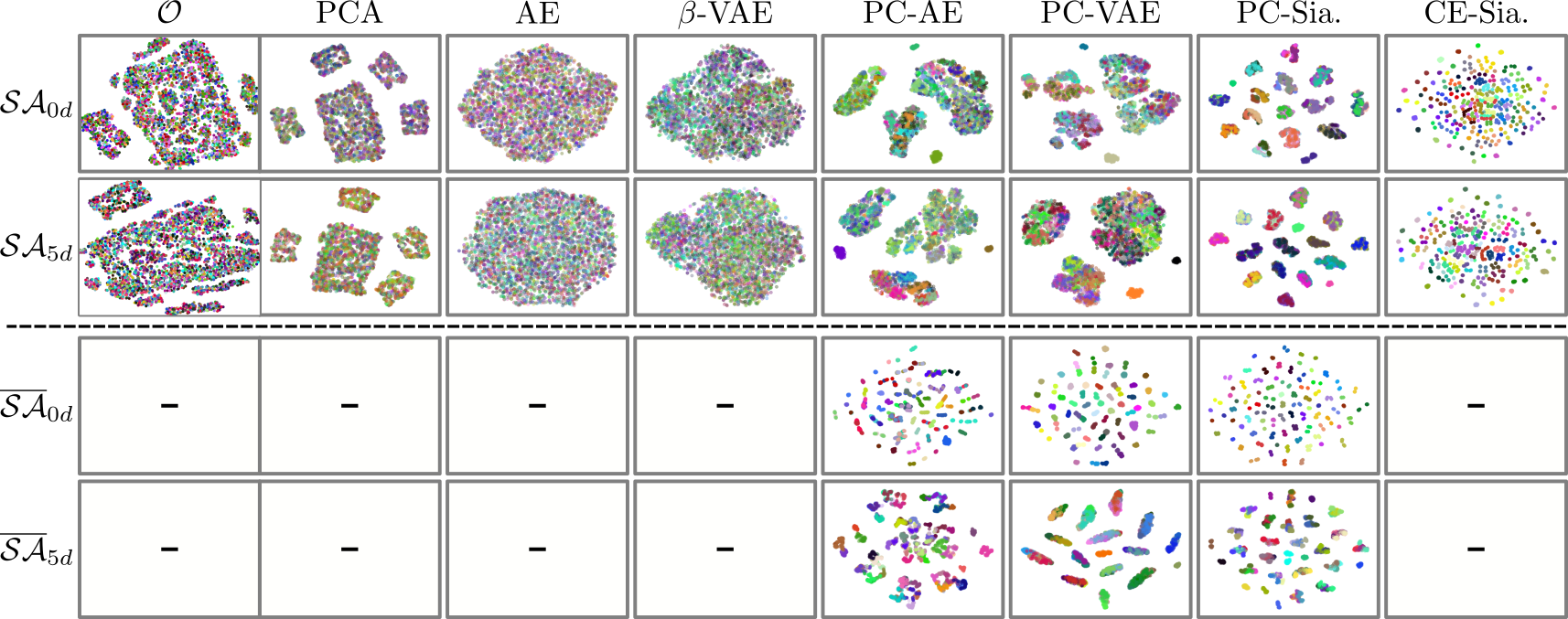}
  \caption{   Two-dimensional t-SNE plots of the latent representations from all models for the shelf arrangement task. All the data augmentations are considered. }
  \label{fig:t-snes_results_shelf_appendix_all}
\end{figure*}

The numbers are easily confirmed with visually inspecting the t-SNE plots in \autoref{fig:t-snes_results_shelf_appendix_all}. We can see that a good separation is only achieved throughout all datasets from the CE-Siamese models.
The random sampling data augmentation in \autoref{sec:problem} helps the models for the no distractor datasets.

\begin{table*}[ht!]
\vspace{-7.3cm}
\resulttable{ $\mathcal{SA}_{0d}$}{
- & $2$ & $0.0$ & $0.33$ & $0.14$ & $2$ & $0.0$ & $0.0$ & $0.0$
\\ \hline
PCA &  $3$ & $0.05$ & $0.29$ & $0.17$ & $5$ & $0.00$ & $0.00$ & $0.00$
\\ \hline 
AE &   $2$ & $0.00$ & $0.38$ & $0.17$ & $2$ & $0.00$ & $0.00$ & $0.00$
\\ \hline 
$ \beta$-VAE  &  $13$ & $0.36$ & $0.56$ & $0.07$ & $11$ & $0.18$ & $0.00$ & $0.00$
\\ \hline 
PC-AE &  $5$ & $0.28$ & \boldmath$1.00$ & $0.93$ & $4$ & $0.75$ & $0.00$ & $0.00$
\\ \hline 
PC-VAE &  $5$ & $0.28$ & \boldmath$1.00$ & $0.78$ & $4$ & $0.75$ & $0.00$ & $0.00$
\\ \hline 
PC-Sia. & \boldmath$16$ & $0.64$ & \boldmath$1.00$ & \boldmath$0.96$ & \boldmath$32$ & $0.59$ & $1.01$ & $1.80$
\\ \hline 
CE-Sia. &  $296$ &\boldmath $1.00$ & \boldmath$1.00$ & $0.54$ & $842$ & \boldmath$1.00$ & \boldmath$95.9$ & \boldmath$95.90$
\\ \hline 
%%%%%%%%%%%%%%%%%%%%% rand 2 act 0 augmentation 
 \hline
&  \multicolumn{8}{c|}{Dataset  $\overline{\mathcal{SA}}_{0d}$}
\\ \hline 
PCA &  - & - & - & - & - & - & - & - %&  $3$ & $0.05$ & $0.29$ & $0.17$ & $5$ & $0.00$ & $0.00$ & $0.00$
\\ \hline 
AE &   - & - & - & - & - & - & - & - % $2$ & $0.01$ & $0.4$ & $-0.04$ & $2$ & $0.5$ & $0.00$ & $0.00$
\\ \hline 
$ \beta$-VAE  &  - & - & - & - & - & - & - & - %&  $2$ & $0.00$ & $0.51$ & $0.06$ & $2$ & $0.00$ & $0.00$ & $0.00$
\\ \hline 
PC-AE &  $97$ & $0.87$ & $0.95$ & $0.51$ & \boldmath$368$ & $0.77$ & $20.30$ & $35.90$
\\ \hline 
PC-VAE &  \boldmath$64$ & $0.90$ & $0.99$ & $0.32$ & $235$ & $0.77$ & $33.13$ & $55.40$
\\ \hline 
PC-Sia. &  $225$ & \boldmath$1.00$ & \boldmath$1.00$ & \boldmath$0.74$ & $772$ & \boldmath$1.00$ & \boldmath$100.0$ & \boldmath$100.0$
\\ \hline 
CE-Sia. &  - & - & - & - & - & - & - & -
\\ \hline 
 }
\resulttable{ $\mathcal{SA}_{5d}$}{
- & $1$ & $0.0$ & $0.36$ & $0.06$ & $0$ & $0.0$ & $0.0$ & $0.0$ \\ \hline
PCA &  $5$ & $0.09$ & $0.30$ & $0.18$ & $9$ & $0.22$ & $0.00$ & $0.00$
\\ \hline 
AE &  $2$ & $0.01$ & $0.43$ & $0.00$ & $2$ & $0.00$ & $0.00$ & $0.00$
\\ \hline 
$ \beta$-VAE  &  $2$ & $0.01$ & $0.52$ & $0.08$ & $2$ & $0.00$ & $0.00$ & $0.00$
\\ \hline 
PC-AE &  $5$ & $0.28$ & \boldmath$1.00$ & $0.9$ & $4$ & $0.75$ & $0.40$ & $0.40$
\\ \hline 
PC-VAE &  $5$ & $0.28$ & \boldmath$1.00$ & $0.79$ & $4$ & $0.75$ & $0.40$ & $0.40$
\\ \hline 
PC-Sia. &  \boldmath$16$ & $0.64$ & \boldmath$1.00$ & \boldmath$0.98$ & \boldmath$32$ & $0.69$ & $2.42$ & $4.10$
\\ \hline 
CE-Sia. &  $286$ & \boldmath$1.00$ & \boldmath$1.00$ & $0.46$ & $841$ & \boldmath$0.99$ & \boldmath$95.12$ & \boldmath$95.30$
\\ \hline
%%%%%%%%%%%%%%%%%%%%% rand 2 act 0 augmentation 
 \hline
&  \multicolumn{8}{c|}{Dataset  $\overline{\mathcal{SA}}_{5d}$} 
\\ \hline
PCA &  - & - & - & - & - & - & - & -
\\ \hline 
AE &  - & - & - & - & - & - & - & -
\\ \hline 
$ \beta$-VAE  &   - & - & - & - & - & - & - & -
\\ \hline 
PC-AE &  $18$ & $0.49$ & $0.98$ & $0.01$ & $34$ & $0.44$ & $0.24$ & $0.40$
\\ \hline 
PC-VAE &  $16$ & $0.64$ & \boldmath$1.00$ & $0.52$ & $32$ & $0.69$ & $2.42$ & $4.10$
\\ \hline 
PC-Sia. &  \boldmath$30$ & \boldmath$0.78$ & \boldmath$1.00$ & \boldmath$0.61$ &\boldmath $87$ & \boldmath$0.74$ & \boldmath$6.66$ & \boldmath$13.10$
\\ \hline 
CE-Sia. &  - & - & - & - & - & - & - & -
\\ \hline 
}
\caption{
Evaluation results of the models for the shelf arrangement task. Datasets  $\mathcal{SA}_{0d}$ (left part) and $\mathcal{SA}_{5d}$ (right part) with respective augmented versions are considered. Best results in bold.  }
\label{tab:results_all_shelf_arranging}
%\vspace{-\baselineskip}
\vspace{-14.5cm}
\end{table*}
%\FloatBarrier

\newpage
\clearpage

\end{document}